%% file: main.tex
\documentclass[10pt,twocolumn,letterpaper]{article}

\usepackage{cvpr}              


\definecolor{cvprblue}{rgb}{0.21,0.49,0.74}
\usepackage[pagebackref,breaklinks,colorlinks,allcolors=cvprblue]{hyperref}

\def\paperID{***} 
\def\confName{CVPR}
\def\confYear{2025}

\usepackage{multirow}
\usepackage{makecell}
\usepackage{bookmark}
\usepackage{algorithm} 
\usepackage{listings}
\usepackage{arydshln}
\usepackage{overpic}

\definecolor{cvprblue}{rgb}{0.21,0.49,0.74}
\definecolor{codeblue}{rgb}{0.25,0.5,0.5}
\definecolor{codekw}{rgb}{0.85, 0.18, 0.50}
\definecolor{std}{rgb}{0.7529,0.4902,0.6471}

\definecolor{a}{rgb}{0.988, 0.784, 0.447}
\definecolor{b}{rgb}{0.961, 0.651, 0.451}
\definecolor{c}{rgb}{0.718, 0.671, 0.686}
\definecolor{d}{rgb}{0.451, 0.420, 0.616}
\definecolor{e}{rgb}{0.925, 0.957, 1.000}

\usepackage{pifont}
\newcommand{\xmark}{\ding{55}}
\newcommand{\cmark}{\ding{51}}
\newcommand{\smark}{\ding{108}}
\newcommand{\rmark}{\ding{93}}

\newcommand{\chunkmark}{\ding{112}}
\newcommand{\arrowmark}{\ding{213}}

\lstdefinestyle{Python}{
  language=python,
  backgroundcolor=\color{white},
  basicstyle=\fontsize{7.5pt}{7.5pt}\ttfamily\selectfont,
  columns=fullflexible,
  breaklines=true,
  captionpos=b,
  commentstyle=\fontsize{7.5pt}{7.5pt}\color{codeblue},
  keywordstyle=\fontsize{7.5pt}{7.5pt}\color{codekw},
  morekeywords={assert},
}

\newcommand{\methodname}{RecConv}
\newcommand{\modelname}{RecNeXt}

\title{\methodname: Efficient Recursive Convolutions for Multi-Frequency Representations}

\author{Mingshu Zhao$^{1}$ \quad Yi Luo$^{1}$ \quad Yong Ouyang$^{1,2}$  \\
$^1$Sichuan Energy Internet Research Institute, Tsinghua University \\
$^2$Chengdu Qingrong Shentong Technology \\
{\tt\small zhaomingshu@tsinghua-eiri.org \quad luoyi@tsinghua-eiri.org \quad ouyangyong@deepsensing.cn}}

\begin{document}
\maketitle
\input{sec/0_abstract}    
\input{sec/1_introduction}
\input{sec/2_related}
\input{sec/3_method}
\input{sec/4_experiment}
\input{sec/5_conclusion}
{
    \small
    \bibliographystyle{ieeenat_fullname}
    \bibliography{main}
}
\input{sec/X_suppl}

\end{document}

%% file: sec/0_abstract.tex
\begin{abstract}
    Recent advances in vision transformers (ViTs) have demonstrated the advantage of global modeling capabilities, prompting widespread integration of large-kernel convolutions for enlarging the effective receptive field (ERF).
    However, the quadratic scaling of parameter count and computational complexity (FLOPs) with respect to kernel size poses significant efficiency and optimization challenges.
    This paper introduces \methodname{}, a recursive decomposition strategy that efficiently constructs multi-frequency representations using small-kernel convolutions. 
    \methodname{} establishes a linear relationship between parameter growth and decomposing levels which determines the effective receptive field $k\times 2^\ell$ for a base kernel $k$ and $\ell$ levels of decomposition, while maintaining constant FLOPs regardless of the ERF expansion. 
    Specifically, \methodname{} achieves a parameter expansion of only $\ell+2$ times and a maximum FLOPs increase of $5/3$ times, compared to the exponential growth ($4^\ell$) of standard and depthwise convolutions.   
    \modelname{}-M3 outperforms RepViT-M1.1 by 1.9 $AP^{box}$ on COCO with similar FLOPs.
    This innovation provides a promising avenue towards designing efficient and compact networks across various modalities.
    Codes and models can be found at \url{https://github.com/suous/RecNeXt}.
\end{abstract}
\vspace{-5mm}

%% file: sec/1_introduction.tex
\section{Introduction}
\label{sec:intro}

\begin{figure}[t]
  \centering
    \begin{overpic}[width=1.0\linewidth]{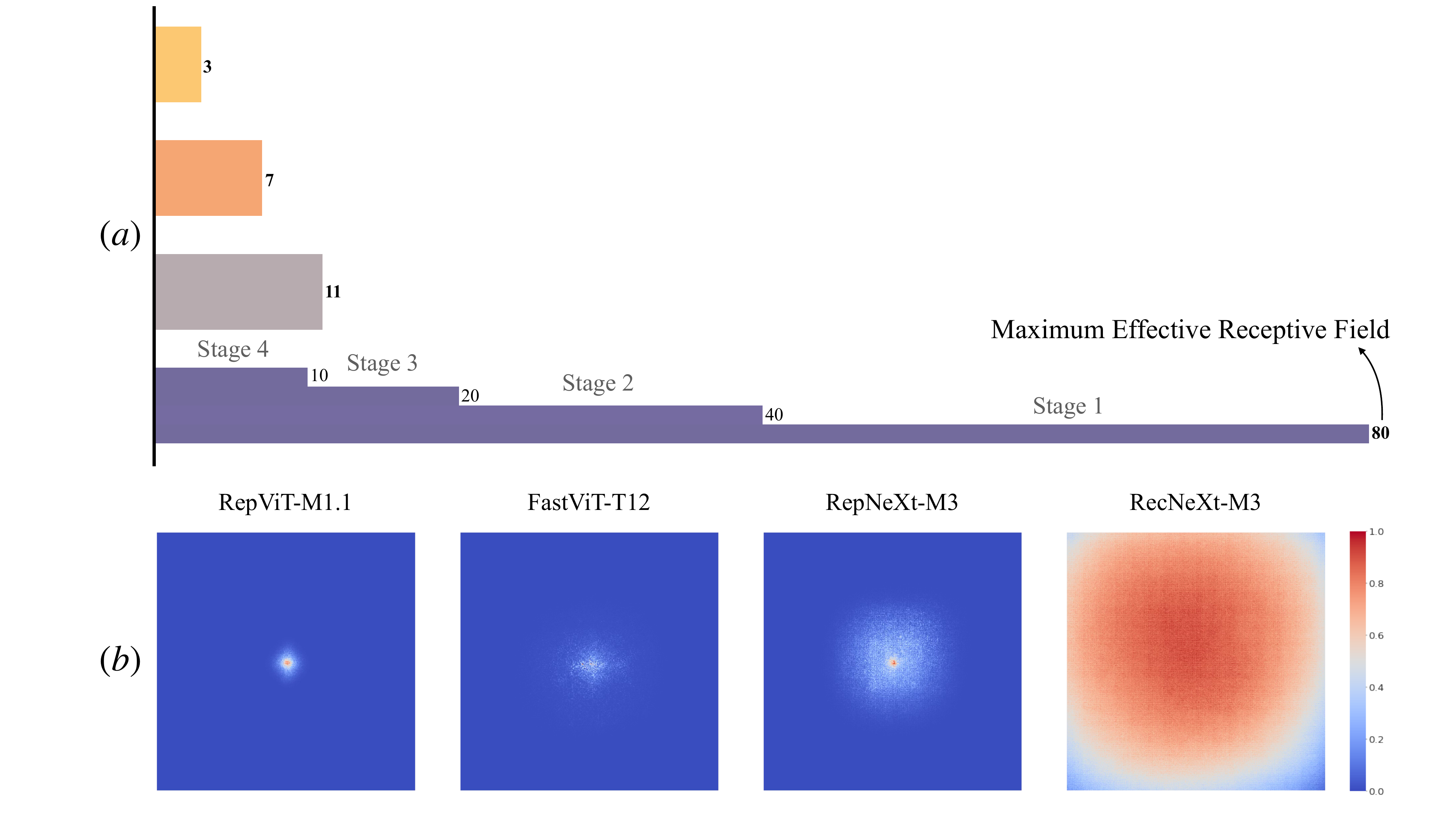}
      \put(25,52){\scalebox{0.49}{\input{tables/performance}}}
    \end{overpic}
    \caption{\textbf{Effective Receptive Field (ERF) and Model Performance}. 
    (a) Comparison of ERFs and performance metrics among \modelname{} and prominent lightweight models.
    Top-1 accuracy is evaluated on ImageNet-1K and the latency is assessed using an iPhone 13 running iOS 18, where all models are formatted into \textit{mlmodel}.
    ``$\dagger$'' denotes pretraining at $256\times256$ resolution and ``$\ast$'' indicates pretraining with hard knowledge distillation.
    (b) The effective receptive fields of RepViT~\cite{wang2023repvit}, FastViT~\cite{vasu2023fastvit}, RepNeXt~\cite{zhao2024repnext}, and \modelname{}. 
    \modelname{} exhibits the broadest ERF, maintaining competitive latency and superior performance without reliance on structural reparameterization (SRP).}
    \label{fig:performance}
    \vspace{-0.5cm}
\end{figure}

Convolutional Neural Networks (CNNs)~\cite{lecun1989backpropagation,krizhevsky2012imagenet,he2016deep,redmon2016you} have been foundational in computer vision for years, exploiting their intrinsic locality and translation equivariance~\cite{dosovitskiy2020image}.
Nonetheless, the recent emergence of vision transformers (ViTs) has demonstrated that establishing global context through self-attention mechanisms can lead to superior performance on various computer vision benchmarks~\cite{dosovitskiy2020image,carion2020end,SETR}.
Following the success of ViTs, extensive research has validated the effectiveness of large kernel convolutions in augmenting model capabilities through expanded receptive fields.
ConvNeXt~\cite{liu2022convnet} and RepLKNet~\cite{ding2022scaling} achieved performance comparable or superior to Swin Transformers~\cite{liu2021swin} when aligned with contemporary design paradigms.
SLaK~\cite{liu2022more} and PeLK~\cite{chen2024pelk} explored kernels beyond $51\times51$ and $101\times101$ through dynamic sparsity and parameter sharing.
Moreover, large-kernel CNNs have shown superiority as teachers in knowledge distillation~\cite{huang2023largekernelsbetterteachers}.

Although large-kernel convolutions provide superior performance, the quadratic scaling of parameter count and computational complexity relative to kernel size presents substantial efficiency challenges and optimization difficulties, restricting their practical deployment on resource-constrained devices.
Consequently, small-kernel convolutions are still preferred~\cite{wang2023repvit,wang2024yolov10,Shaker_2023_ICCV,li2022rethinking} in such scenarios considering their optimized algorithms and compact weights.
Recent studies have explored large-kernel convolutions in resource-constrained environments.
FastViT~\cite{vasu2023fastvit}, StarNet~\cite{ma2024rewrite}, and EfficientMod~\cite{ma2024efficient} employed $7\times7$ depthwise convolutions to capture long-range dependencies.
EdgeNeXt~\cite{Maaz2022EdgeNeXt} introduced adaptive kernel mechanisms for multi-level feature extraction.
MobileMamba~\cite{mobilemamba} enhanced multi-scale representations through a multi-receptive field feature interaction module.
RepNeXt~\cite{zhao2024repnext} employed multi-branch efficient operations to replace the $11\times11$ depthwise convolution.
However, the application of even larger convolution kernels in resource-constrained settings remains unexplored.

OctConv~\cite{Chen:ICCV:2019:octave} leveraged octave convolutions to enable multi-frequency feature representation.
WaveMix~\cite{jeevan2023wavemix} and WTConv~\cite{finder2024wavelet} incorporated small-kernel convolutions with wavelet decomposition to obtain global receptive fields.
HorNet~\cite{rao2022hornet} performed high-order spatial interactions with gated convolutions and recursive designs.
While effective, their implementation requires substantial amount of parameters and resource-intensive operations.
Motivated by their multi-frequency response, we propose a recursively decomposed convolution (\methodname{}) optimized for resource-constrained mobile vision applications.
Based on \methodname{}, we introduce \modelname{}, a simple yet effective vision backbone that delivers competitive efficiency with superior performance.
Our contributions can be concluded as follows:
\begin{itemize}[leftmargin=1em]
\setlength
\itemsep{0em}
    \item Inspired by WTConv~\cite{finder2024wavelet}, we develop a recursive decomposition strategy to generate multi-frequency representations. 
    This approach establishes a linear relationship between parameter growth and decomposing levels, while maintaining constant FLOPs, making it highly adaptable for designing efficient networks with large receptive fields across diverse modalities.
    \item Following this design principle, we construct \methodname{} as a plug-and-play module that can be seamlessly integrated into existing computer vision architectures.
    \item Leveraging \methodname{}, we introduce \modelname{}, two series of efficient and effective vision backbones equipped with large receptive fields and optimized towards resource-constrained mobile vision applications.
    \item Extensive experiments across various vision benchmarks demonstrate the flexibility and effectiveness of \modelname{} without relying on neural architecture search (NAS) or structural reparameterization (SRP).
\end{itemize}

%% file: tables/performance.tex
  \begin{tabular}{cccccccccc}
  \toprule
   & \multirow{2}{*}{\makecell[c]{Model}}         &  Latency & MACs & ImageNet-1K & COCO & ADE20K & \multirow{2}{*}{\makecell[c]{SRP}}
  \\
  & & (ms) & (G) & Top-1(\%) & AP$^{box}$ & mIoU & 
  \\
  \hline
  \textcolor{a} {\smark} & RepViT-M1.1 & 1.5 & 1.3 & 79.4 / 80.7$^\ast$ & 39.8 & 40.6 & \cmark \\
  \textcolor{b} {\smark} & FastViT-T12\textsuperscript{$\dagger$} & 1.6 & 1.4 & 79.1 / 80.3$^\ast$ & - & - & \cmark \\
  \textcolor{c} {\smark} & RepNeXt-M3 & 1.5 & 1.3 & 79.4 / 80.7$^\ast$ & 40.8 & 40.6 & \cmark \\
  \rowcolor[HTML]{ECF4FF}
  \textcolor{d} {\smark} & \modelname{}-M3 & 1.6 & 1.4 & \text{79.6} / \text{80.9}$^\ast$ & \text{41.7} & \text{41.0} & \xmark \\
  \rowcolor[HTML]{ECF4FF}
  \textcolor{d} {\rmark} & \modelname{}-A3 & 2.4 & 1.4 & \textbf{80.1} / \textbf{81.1}$^\ast$ & \textbf{42.1} & \textbf{41.9} & \xmark \\
  \bottomrule
  \end{tabular}

%% file: sec/2_related.tex
\section{Related Work}

\noindent\textbf{Large Kernel CNNs}: 
Large kernels have been explored in early CNNs~\cite{krizhevsky2012imagenet,szegedy2015going}, but fell out of favor after VGG~\cite{simonyan2014very}.
ConvNeXt~\cite{liu2022convnet} and ConvMixer~\cite{convmixer} embraced contemporary design philosophy with large-kernel depthwise convolutions, replicating the global perspective of ViTs~\cite{dosovitskiy2020image} and MLP-Mixers~\cite{tolstikhin2021mlp}.
MogaNet~\cite{iclr2024MogaNet} incorporated multi-scale spatial aggregation blocks with dilated convolutions to capture discriminative features.
SKNet~\cite{li2019selective} and LSKNet~\cite{Li_2023_ICCV} employed multi-branch selective mechanisms along the channel or spatial dimension.
RepLKNet~\cite{ding2022scaling} achieved performance comparable to Swin Transformers~\cite{liu2021swin} by expanding kernel size to $31\times31$ with SRP.
UniRepLKNet~\cite{ding2023unireplknet} demonstrated the universal applicability of large-kernel designs across various modalities.
SLaK~\cite{liu2022more} improved performance by expanding kernels beyond $51\times51$ using stripe convolutions and dynamic sparsity.
PeLK~\cite{chen2024pelk} explored kernels up to $101\times101$ in a human-like pattern without performance saturation.
GFNet~\cite{Rao:NIPS2021:gfnet} leveraged Fourier transforms to capture long-term spatial dependencies in the frequency domain. 
WTConv~\cite{finder2024wavelet} replaced large-kernel convolutions with a set of small-kernel convolutions through wavelet decomposition.
Furthermore, $n\times n$ convolutions are often factorized into sequential or parallel $1\times n$ and $n\times 1$ convolutions for efficiency~\cite{szegedy2016rethinking,szegedy2017inception,peng2017large,guo2022segnext,yu2023inceptionnext}.
Additionally, LargeKernel3D~\cite{chen2023largekernel3d} and ModernTCN~\cite{donghao2024moderntcn} extended large kernel design to 3D networks and time series analysis.

\noindent\textbf{Efficient Networks}: 
Developing efficient networks for resource-constrained vision applications has attracted significant attention~\cite{howard2017mobilenets,redmon2016you,wang2024yolov10}.
MobileNets~\cite{howard2017mobilenets,sandler2018mobilenetv2,howard2019searching} proposed depthwise separable convolutions and inverted residual blocks to optimize the efficiency-accuracy trade-off.
GhostNet~\cite{ghostnet}, FasterNet~\cite{FasterNet}, and SHViT~\cite{SHVIT} employed partial channel operations for parameter reduction and computational efficiency.
MobileOne~\cite{vasu2023mobileone}, FastViT~\cite{vasu2023fastvit}, and RepViT~\cite{wang2023repvit} utilized structural reparameterization (SRP) to improve feature diversity without increasing the inference latency.
MnasNet~\cite{tan2019mnasnet} and EfficientNet~\cite{tan2019efficientnet} leveraged neural architecture search (NAS) to automatically discover efficient architectures.
MobileViTs~\cite{mehta2021mobilevit,mehta2022separable,wadekar2022mobilevitv3}, EdgeViTs~\cite{pan2022edgevits}, and EfficientFormers~\cite{li2022efficientformer,li2022rethinking} designed hybrid architectures to balance performance and efficiency.
Mobile-Former~\cite{chen2022mobile} and ParC-Net~\cite{zhang2022parcnet} introduced multi-branch feature integration to improve representational capacity.
SwiftFormer~\cite{Shaker_2023_ICCV} and LightViT~\cite{huang2022lightvit} constructed linear attention mechanisms to replace quadratic matrix multiplications.
FastViT~\cite{vasu2023fastvit}, StarNet~\cite{ma2024rewrite}, and EfficientMod~\cite{ma2024efficient} adopted $7\times7$ depthwise convolutions to capture long-range dependencies.
EdgeNeXt~\cite{Maaz2022EdgeNeXt} and RepNeXt~\cite{zhao2024repnext} introduced multi-branch efficient operations to obtain large-kernel receptive fields.
MobileMamba~\cite{mobilemamba} incorporated long-range wavelet transform-enhanced mamba and efficient multi-kernel depthwise convolution to enhance multi-scale perception.

The major perspective of our work is inspired by WTConv~\cite{finder2024wavelet}, while our design focuses on resource-constrained applications.
And there are four major differences between WTConv and our proposed method:
(1) We adopt a shared and learnable depthwise convolutional downsampling layer with a stride of 2.
(2) We maintain the same channel dimension across different scales for structural consistency and computational efficiency.
(3) We use bilinear interpolation (interchangeable during inference) to reduce parameter count and memory footprint.
(4) Our design supports simultaneous recursive decomposition along both spatial and channel dimensions.

%% file: sec/3_method.tex
\section{Method}

\subsection{Overall Architecture}
\label{sec:architecture}

\begin{table*}[t]
  \caption{\textbf{Classification performance on ImageNet-1K.} Following~\cite{li2022efficientformer,wang2023repvit}, NPU latency is measured on an iPhone 13 with models compiled by Core ML Tools.
  CPU latency is accessed on a Quad-core ARM Cortex-A57 processor.
  Throughput is tested on an Nvidia RTX3090 GPU with maximum power-of-two batch size that fits in memory. ``$\dagger$'' denotes the evaluation image size is 256.
  Gray annotations indicate replacing bilinear interpolation with nearest interpolation during inference, improving GPU throughput at a slight accuracy cost.
  }
  \vspace{-4pt}
  \label{tab:with_distillation}
  \centering
  \small
  \scalebox{0.95}{\input{tables/classification_with_distillation}}
  \vspace{-15pt}
\end{table*}

\begin{figure}[t]
\centering
    \includegraphics[width=1.0\linewidth]{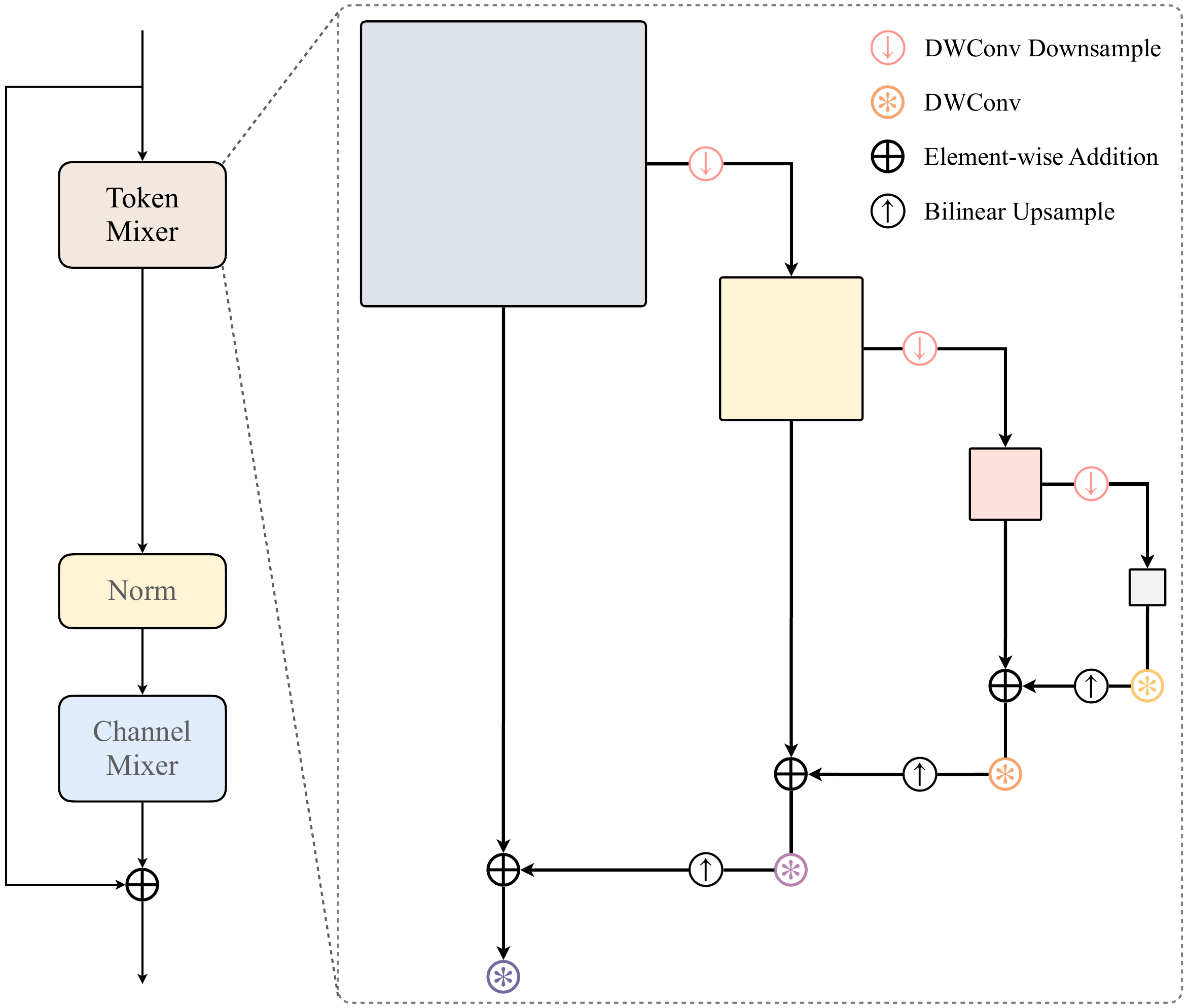}
    \caption{
        \textbf{Recursive Convolution Module (\methodname{})}.
        The left side illustrates a MetaNeXt block abstracted from previous works~\cite{yu2022metaformer,liu2022convnet,yu2023inceptionnext}.
        The right side highlights the \methodname{} details, where feature maps are recursively decomposed into multiple spatial scales using a shared depthwise convolutional downsampling layer.
        Each scale applies a small-kernel depthwise convolution to generate features of varying spatial sizes, which are then upsampled and aggregated, collectively replicating the large-kernel receptive field while preserving computational and parameter efficiency.
        This design strategy accommodates various functions and can be extended to other modalities.
    }
    \label{fig:architecture}
    \vspace{-15pt}
\end{figure}

The \modelname{} architecture builds upon RepViT~\cite{wang2023repvit} and RepNeXt~\cite{zhao2024repnext}, adhering to the four-stage framework of conventional CNNs~\cite{he2016deep} and hierarchical ViTs~\cite{liu2021swin}.
Starting with two $3\times3$ convolutions with a stride of $2$~\cite{wang2023repvit,li2022rethinking,xiao2021early}, progressively enhancing semantic representations while reducing spatial dimensions. 
The micro-blocks adopt the MetaNeXt design~\cite{liu2022convnet,yu2023inceptionnext}, incorporating a \textit{token mixer} for spatial feature extraction, a \textit{channel mixer} for visual semantic interaction, a normalization layer~\cite{batch_norm, layer_norm} for training stability, and a shortcut connection~\cite{he2016deep} to facilitate smoother optimization~\cite{visualloss}.
\begin{equation}
  Y = X + \mathrm{ChannelMixer}\big(\mathrm{Norm}(\mathrm{TokenMixer}(X))\big),
\end{equation}
where $X, Y \in \mathbb{R}^{B \times C \times H \times W}$ with $B$ represents batch size, $C$ denotes channel number, and $H$ and $W$ indicate image height and width, respectively. 
$\mathrm{Norm}(\cdot)$ denotes the Batch Normalization (BN) layer~\cite{batch_norm}, which can be fused into the previous convolution layer.
$\mathrm{TokenMixer}(\cdot)$ implements the \methodname{} and $\mathrm{ChannelMixer}(X)=\mathrm{Conv_{1\times1,\downarrow}}\big(\sigma(\mathrm{Conv_{1\times1,\uparrow}}(X))\big)$ is a channel MLP module comprising two fully-connected layers with a GELU~\cite{hendrycks2016gelu} activate function $\sigma$ in between, resembling the feed-forward network in a Transformer~\cite{vaswani2017attention}. 
$\mathrm{Conv_{1\times1,\uparrow}}$ and $\mathrm{Conv_{1\times1,\downarrow}}$ stand for $1\times1$ pointwise convolutions for expanding and squeezing feature maps, respectively.

The downsampling layer between each stage is a modified version of the MetaNeXt block~\cite{liu2022convnet,yu2023inceptionnext}, where the shortcut connection bypasses the \textit{channel mixer}.
\begin{equation}
\begin{split}
  \hat{X} &= \mathrm{Norm}(\mathrm{TokenMixer}(X)), \\ 
  Y &= \hat{X} + \mathrm{ChannelMixer}(\hat{X}),
\end{split}
\end{equation}
where $\hat{X}, Y \in \mathbb{R}^{B \times 2C \times H/2 \times W/2}$. 
$\mathrm{TokenMixer}(\cdot)$ is a $7\times7$ depthwise convolution with stride $2$ for downsampling.

\subsection{Recursive Convolution}

The right side of Figure~\ref{fig:architecture} spotlights the \methodname{} details, where initial feature maps are recursively decomposed into multiple spatial scales via a shared depthwise convolutional downsampling layer.
Each scale employs a small-kernel depthwise convolution to extract features at varying spatial resolutions, which are then progressively fused back into the original feature map using bilinear upsampling and element-wise addition for simplicity.
This efficient and recursive design enables the model to encapsulate multi-frequency representations within each block, effectively replicating the receptive field of large-kernel convolutions without the associated parameter inflation and computational overhead.
Furthermore, this flexible and modular design philosophy is compatible with diverse operators, and can be seamlessly integrated into existing vision architectures or extended to other modalities.

\begin{algorithm}[t]
\caption{\methodname{} in a PyTorch-like style}
\label{alg:recconv}
\begin{lstlisting}[style=Python]
class RecConv(Module):
  def __init__(self, ic, ks=5, level=1):
      super().__init__()
      self.level = level
      kwargs = {
        'in_channels': ic,
        'out_channels': ic,
        'kernel_size': ks,
        'padding': ks // 2,
        'groups': ic
      }
      # downsample with shared parameters
      self.down = Conv(stride=2, **kwargs)
      # convolutions on different levels
      self.conv = [Conv(**kwargs)] * (level+1)

  def forward(self, x):
      i, fs = x, []
      for _ in range(self.level):
          x, s = self.down(x), x.shape[2:]
          fs.append((x, s))

      x = 0
      for c, (f, s) in zip(self.conv, reversed(fs)):
          x = interpolate(c(f + x), size=s)
      return self.conv[self.level](i + x)
\end{lstlisting}
\end{algorithm}

Specifically, as demonstrated in Algorithm~\ref{alg:recconv} and Equation~\ref{eq:recconv}, feature maps of level $\ell$ are produced by the previous level $\ell-1$ using a depthwise convolution with a stride of $2$.
This process then generates new feature maps at level $\ell+1$ using the same downsampling operation.
Higher-level feature maps are processed through depthwise convolutions and then progressively integrated back into previous levels through bilinear interpolation and element-wise addition.
This recursive design maintains parameter efficiency as decomposition levels increase by reusing the downsampling layer and employing a parameter-free upsampling operation across all decomposition levels.
\begin{equation}
  \label{eq:recconv}
  \begin{split}
    X_{\ell} &= \mathrm{DWConv_{k\times k, \downarrow}}(X_{\ell-1}), \\
    X_{\ell} &= X_{\ell} + \mathrm{Interp_{\uparrow}}\left(\mathrm{DWConv_{k\times k}^{(\ell+1)}}(X_{\ell+1})\right),
  \end{split}
\end{equation}
where $X_{\ell-1} \in \mathbb{R}^{B \times C \times 2h \times 2w}$, $X_{\ell} \in \mathbb{R}^{B \times C \times h \times w}$, and $X_{\ell+1} \in \mathbb{R}^{B \times C \times \frac{h}{2} \times \frac{w}{2}}$ represent feature maps at levels $\ell-1$, $\ell$, and $\ell+1$, respectively.
$\mathrm{DWConv_{k\times k, \downarrow}}$ indicates the shared-weight depthwise convolution with a kernel size of $k\times k$ and a stride of $2$, while $\mathrm{DWConv_{k\times k}^{(\ell+1)}}$ denotes the depthwise convolution with a kernel size of $k\times k$ at level $\ell+1$.
$\mathrm{Interp_{\uparrow}}$ stands for bilinear interpolation (interchangeable during inference) upsampling operation. 
Features at different scales are progressively fused back into the original feature map, effectively replicating multi-frequency representations with large-kernel receptive fields.

\begin{table}[t]
\caption{\textbf{Complexity of different types of convolution.} 
  The measurement is streamlined by focusing on convolution operations, with the assumption of consistent input and output channels, and excluding the bias term.
  $k$, $C$, $H$ and $W$ denote the base kernel size, channel number, image height and width, respectively.
  And $\ell$ represents the number of decomposition levels that determine the effective receptive field, which is $k\times 2^\ell$.
}
\label{tab:complexity}
  \centering
  \small
  \renewcommand{\arraystretch}{1.1}
  \begin{tabular}{clll}
  \toprule
  \multirow{1}{*}{Convolution}          &  \multirow{1}{*}{\makecell{Parameters}} & \multirow{1}{*}{FLOPs} \\
  \hline
  Standard & $\color{codeblue}{k^2C}$$\times (4^\ell C)$ & $\color{codeblue}{k^2CHW}$$\times (4^\ell C)$ \\
  Depthwise & $\color{codeblue}{k^2C}$$\times (4^\ell)$ & $\color{codeblue}{k^2CHW}$$\times (4^\ell)$ \\
  RecConv & $\color{codeblue}{k^2C}$$\times (\ell+2)$ & $\color{codeblue}{k^2CHW}$$\times (5/3)$ \\
  \bottomrule
  \end{tabular}
  \vspace{-10pt}
\end{table}

Table~\ref{tab:complexity} illustrates that to achieve a larger effective kernel size of $k\times 2^\ell$, both standard and depthwise convolutions require exponential growth in parameters and FLOPs with respect to the decomposition level $\ell$.
In contrast, \methodname{} exhibits only linear growth in parameters while maintaining constant FLOPs.
Specifically, \methodname{} consists of a shared-weight depthwise convolutional downsampling module and $\ell+1$ depthwise convolution layers, which primarily contribute to the parameter count and FLOPs.
The parameter count remains consistent across the downsampling layer and depthwise convolutions at each level, growing by a factor of $\ell+2$ compared to the base kernel.
FLOPs can be regarded as a series of geometric progression, decreasing exponentially as the levels increase, as demonstrated in Equation~\ref{eq:flops}.
Consequently, \methodname{} achieves linear parameter growth and nearly constant FLOPs proportional to the decomposition level, making it an efficient solution for attaining large receptive fields in resource-constrained scenarios.
\begin{equation}
  \label{eq:flops}
  1+2\sum_{n=1}^{\ell}\frac{1}{4^n} < \sum_{n=0}^{\infty}2\cdot \left(\frac{1}{4}\right)^n - 1 = \frac{5}{3} 
\end{equation}

In addition, our \methodname{} embodies a versatile design philosophy that accommodates different operators and dimensions.
For instance, the shared downsampling layer can be substituted with a sequence of channel-wise chunking followed by depthwise convolutions; 
depthwise convolutions at each level can be replaced by efficient attention mechanisms~\cite{han2023flatten,han2024demystify}; 
bilinear upsampling operations can be converted to transposed convolutions;
element-wise addition can be reconfigured as channel-wise concatenation. 

%% file: tables/classification_with_distillation.tex
\begin{tabular}{ccccccccccc} \toprule
    Model           & Type      & Params & MACs &  \multicolumn{2}{c}{Latency (ms) $\downarrow$} & Throughput $\uparrow$ & Top1 $\uparrow$ & \multicolumn{2}{c}{Auxiliary}
    \\
    & & (M) & (G) & NPU & CPU & (im/s) & (\%) & NAS & SRP
    \\
    \hline
    MobileViG-Ti~\cite{munir2023mobilevig}  &   CNN-GNN   & 5.2      &  0.7  & 1.4  & 261    & 4337   & 75.7 & \xmark & \xmark \\
    SwiftFormer-XS~\cite{Shaker_2023_ICCV} & Hybrid & 3.5 & 0.6 & 1.5 & 194  & 4304  & 75.7 & \xmark & \xmark \\
    EfficientFormerV2-S0~\cite{li2022rethinking}  &   Hybrid   & 3.5      &  0.4  & 1.4 & 198        & 1274  & 75.7  & \cmark & \xmark \\
    FastViT-T8\textsuperscript{$\dagger$}~\cite{vasu2023fastvit} & Hybrid & 3.6 & 0.7 & 1.3 & 280  & 3909  & 76.7 & \xmark & \cmark \\
    RepViT-M0.9~\cite{wang2023repvit} & CONV & 5.1 & 0.8 & 1.3 & 222 & 4817  & 78.7  & \xmark & \cmark \\
    RepNeXt-M1~\cite{zhao2024repnext} & CONV & 4.8 & 0.8 & 1.3 & 267 & 3885   & 78.8  & \xmark & \cmark \\
    EfficientFormerV2-S1~\cite{li2022rethinking}  &   Hybrid   & 6.1      &  0.7 & 1.6 & 303        & 1153   & 79.0  & \cmark & \xmark \\
    RepViT-M1.0~\cite{wang2023repvit} & CONV & 6.8 & 1.1 & 1.4  & 280 & 3910  & 80.0  & \xmark & \cmark \\
    RepNeXt-M2~\cite{zhao2024repnext} & CONV & 6.5 & 1.1 & 1.4  & 325 & 3198 & 80.1  & \xmark & \cmark \\
    \rowcolor[HTML]{ECF4FF}
    \textbf{\modelname{}-M1} & CONV & 5.2 & 0.9 & 1.4 & 361 & 384 / \textcolor{gray}{3303} & \textbf{79.2} / \textcolor{gray}{78.8}  & \xmark & \xmark \\
    \rowcolor[HTML]{DAE8FC}
    \textbf{\modelname{}-M2} & CONV & 6.8 & 1.2 & 1.5 & 431 & 325 / \textcolor{gray}{2724} & \textbf{80.3} / \textcolor{gray}{80.1} & \xmark & \xmark \\
    \hline
    MobileViG-S~\cite{munir2023mobilevig}  &   CNN-GNN   & 7.2      &  1.0 & 1.6  & 383       & 2985   & 78.2    & \xmark & \xmark \\
    SwiftFormer-S~\cite{Shaker_2023_ICCV} & Hybrid & 6.1 & 1.0 & 1.8 & 278  & 3376  & 78.5 & \xmark & \xmark \\
    EfficientFormer-L1~\cite{li2022efficientformer}       & Hybrid    &    12.3   & 1.3 & 1.8 & 299      & 3360  & 79.2    & \cmark & \xmark \\
    FastViT-T12\textsuperscript{$\dagger$}~\cite{vasu2023fastvit} & Hybrid & 6.8 & 1.4 & 1.8 & 456  & 3182  & 80.3 & \xmark & \cmark \\
    RepViT-M1.1~\cite{wang2023repvit} & CONV & 8.2 & 1.3 & 1.5 & 326 & 3604  & 80.7 & \xmark & \cmark \\
    RepNeXt-M3~\cite{zhao2024repnext} & CONV & 7.8 & 1.3 & 1.5 & 373 & 2903  & 80.7 & \xmark & \cmark \\
    \rowcolor[HTML]{DAE8FC}
    \textbf{\modelname{}-M3} & CONV & 8.2 & 1.4 & 1.6 & 482 & 314 / \textcolor{gray}{2514} & \textbf{80.9} / \textcolor{gray}{80.5}& \xmark & \xmark \\
    \hline
    MobileViG-M~\cite{munir2023mobilevig}  &   CNN-GNN   & 14.0      &  1.5 & 2.0  & 488       & 2491   & 80.6    & \xmark & \xmark \\
    FastViT-S12\textsuperscript{$\dagger$}~\cite{vasu2023fastvit} & Hybrid & 8.8 & 1.8 & 2.0 & 549  & 2313  & 80.9 & \xmark & \cmark \\
    SwiftFormer-L1~\cite{Shaker_2023_ICCV} & Hybrid & 12.1 & 1.6 & 2.2  & 411 & 2576  & 80.9 & \xmark & \xmark \\
    EfficientFormerV2-S2~\cite{li2022rethinking}  &   Hybrid   &  12.6  &  1.3 & 2.3 & 498        & 611  & 81.6  & \cmark & \xmark \\
    FastViT-SA12\textsuperscript{$\dagger$}~\cite{vasu2023fastvit} & Hybrid & 10.9 & 1.9 & 2.3 & 569  & 2181  & 81.9 & \xmark & \cmark \\
    RepViT-M1.5~\cite{wang2023repvit} & CONV & 14.0 & 2.3 & 1.9 & 540 & 2151  & 82.3 & \xmark & \cmark \\
    RepNeXt-M4~\cite{zhao2024repnext} & CONV & 13.3 & 2.3 & 1.9 & 632 & 1745  & 82.3  & \xmark & \cmark \\
    \rowcolor[HTML]{DAE8FC}
    \textbf{\modelname{}-M4} & CONV & 14.1 & 2.4 & 2.4 & 843  & 169 / \textcolor{gray}{1495} & \textbf{82.5} / \textcolor{gray}{82.0} & \xmark & \xmark \\
    \hline
    EfficientFormer-L3~\cite{li2022efficientformer}  &   Hybrid   & 31.3   &  3.9 & 3.4  & 795       & 1422  & 82.4  & \cmark & \xmark \\
    MobileViG-B~\cite{munir2023mobilevig} & CNN-GNN & 26.7 & 2.8 & 3.5 & 865 & 1446  & 82.6 & \xmark & \xmark \\
    SwiftFormer-L3~\cite{Shaker_2023_ICCV} & Hybrid & 28.5 & 4.0 & 3.4 & 859 & 1474  & 83.0 & \xmark & \xmark \\
    EfficientFormer-L7~\cite{li2022efficientformer}  &   Hybrid   & 82.1   &  10.2 & 7.1 & 1906     & 619  & 83.3  & \cmark & \xmark \\
    EfficientFormerV2-L~\cite{li2022rethinking}  &   Hybrid   & 26.1   &  2.6 & 3.8  & 877       & 399  & 83.3  & \cmark & \xmark \\
    RepViT-M2.3~\cite{wang2023repvit} & CONV & 22.9 & 4.5 & 2.8 & 1007 & 1184  & 83.3  & \xmark & \cmark \\
    FastViT-SA24\textsuperscript{$\dagger$}~\cite{vasu2023fastvit} & Hybrid & 20.6 & 3.8 & 3.4  & 1044   & 1128  & \textbf{83.4} & \xmark & \cmark \\
    RepNeXt-M5~\cite{zhao2024repnext} & CONV & 21.7 & 4.5 & 2.8 & 1144 & 978   & 83.3  & \xmark & \cmark \\
    \rowcolor[HTML]{DAE8FC}
    \textbf{\modelname{}-M5} & CONV & 22.9 & 4.7 & 3.4 & 1487 & 104 / \textcolor{gray}{847} & 83.3 / \textcolor{gray}{83.0} & \xmark & \xmark \\
    \bottomrule
    \end{tabular}

%% file: sec/4_experiment.tex
\section{Experiments}

We demonstrate  the versatility and efficacy of \modelname{} across multiple vision tasks: classification on ImageNet-1K~\cite{deng2009imagenet}, object detection and instance segmentation on MS-COCO 2017~\cite{lin2014microsoft}, and semantic segmentation on ADE20K~\cite{zhou2017scene}. 
Following prior works~\cite{li2022efficientformer,li2022rethinking,vasu2023mobileone,mehta2022separable,wang2023repvit,zhao2024repnext}, we export the model using Core ML Tools~\cite{coreml} and assess its latency on an iPhone 13 (iOS 18) using the Xcode performance tool. 
Additionally, we measure throughput on an Nvidia RTX3090 GPU with the maximum power-of-two batch size that can be accommodated in memory.

\subsection{Image Classification}

\paragraph{Implementation details.} We perform image classification on ImageNet-1K using a standard image resolution of 224$\times$224 for both training and evaluation. 
This dataset consists of 1.3M training, 50k validation, and 100k test images across 1000 categories.
All models are trained from scratch for 300 epochs following the training protocol in~\cite{vasu2023fastvit,li2022rethinking,wang2023repvit,zhao2024repnext}.
For fair comparisons, we employ the RegNetY-16GF~\cite{radosavovic2020designing} model (top-1 accuracy 82.9\%) as the teacher model for hard knowledge distillation. 
Latency is measured on an iPhone 13 with models compiled via Core ML Tools at a batch size of 1.
Performance with and without distillation is reported in Table~\ref{tab:with_distillation} and~\ref{tab:without_distillation}, respectively.

\begin{table}[t]
  \caption{Results on ImageNet-1K without distillation, where``$\dagger$'' denotes evaluation at $256\times256$ image size.
  \modelname{} achieves higher accuracy on smaller models, but experiences performance saturation with larger variants.
  }
  \vspace{-4pt}
  \label{tab:without_distillation}
  \centering
  \small
  \scalebox{0.85}{\input{tables/classification_without_distillation}}
  \vspace{-15pt}
\end{table}

\begin{table*}[h]
  \centering
  \small
  \caption{
  \textbf{Object detection and instance segmentation} were evaluated using Mask R-CNN on MS-COCO 2017. 
  \textbf{Semantic segmentation} results were obtained on ADE20K. Backbone latencies were measured on an iPhone 13 with 512$\times$512 image crops using Core ML Tools. Models marked with ``$\dagger$'' were initialized with weights pretrained for 450 epochs on ImageNet-1K.
  }
  \vspace{-4pt}
  \resizebox{0.9\linewidth}{!}{\input{tables/object_detection}}
  \vspace{-4pt}
  \label{tab:coco}
\end{table*}

\paragraph{Results with knowledge distillation.}
As demonstrated in Table~\ref{tab:with_distillation}, \modelname{} consistently delivers superior performance and competitive latency across various model sizes without requiring additional training auxiliaries.
Notably, smaller \modelname{} models outperform their RepNeXt counterparts by 0.2\% in top-1 accuracy while maintaining comparable latency and FLOPs.
\modelname{}-M5 matches the top-1 accuracy (83.3\%) of EfficientFormerV2-L with fewer parameters and lower latency.
Without leveraging structural reparameterization (SRP) or neural architecture search (NAS), the \modelname{} series consistently surpasses SwiftFormer in top-1 accuracy with similar computational complexity.
However, \modelname{} encounters challenges with throughput due to the extensive employment of bilinear upsampling operations, which can be replaced by nearest interpolation during inference.
And can also be alleviated by leveraging transposed depthwise convolutions at the cost of increased computational complexity and parameter count.

\paragraph{Results without knowledge distillation.} 
Table~\ref{tab:without_distillation} illustrates that \modelname{} maintains strong performance across various model scales. 
\modelname{}-M1 achieves a top-1 accuracy of 78.0\% with a latency of 1.4 ms and 5.2M parameters, outperforming prior lightweight models such as RepViT-M0.9 (77.4\%) and RepNeXt-M1 (77.5\%) at comparable latencies.
\modelname{}-M2 further improves to 79.2\% accuracy with 6.8M parameters and 1.5 ms latency, surpassing RepViT-M1.0 (78.6\%) and RepNeXt-M2 (78.9\%) with similar FLOPs.
For higher-capacity models, \modelname{}-M3 achieves a top-1 accuracy of 79.6\%, exceeding other leading models without relying on structural reparameterization (SRP) and neural architecture search (NAS).
However, larger models like \modelname{}-M4 and \modelname{}-M5 exhibit performance saturation alongside increased latency, with \modelname{}-M5 plateaus at 81.6\% top-1 accuracy but with a latency of 3.4ms (0.6ms higher than RepNext-M5).
Although \modelname{}-M5 experiences performance degradation, it still outperforms PoolFormer-S36 (81.4\%) with fewer parameters and lower latency.
Additionally, incorporating SRP, carefully tuned configurations, or advanced operators could potentially mitigate this limitation.

\subsection{Downstream Tasks}
\vspace{-4pt}

\paragraph{Object Detection and Instance Segmentation.}
We evaluate \modelname{}'s transfer ability on object detection and instance segmentation tasks.
Following \cite{li2022rethinking}, we integrate \modelname{} into the Mask-RCNN framework~\cite{he2017mask} and conduct experiments on the MS-COCO 2017 dataset~\cite{lin2014microsoft}, presenting metrics of AP$^{box}$ and AP$^{mask}$.
As shown in \Cref{tab:coco}, \modelname{} demonstrates superior performance in terms of AP$^{box}$ and AP$^{mask}$ while maintaining competitive latency. 
For instance, \modelname{}-M3 achieves a notable AP$^{box}$ of 41.7 and AP$^{mask}$ of 38.6 while maintaining a latency of 5.2ms, offering an optimal trade-off between speed and precision.
\modelname{}-M4 surpasses FastViT-SA24 by 1.5 AP$^{box}$ and 1.7 AP$^{mask}$ with 4ms latency improvement, and increases the AP$^{box}$ and AP$^{mask}$ by 0.6 compared to RepNext-M4 with only 0.5ms latency increase.
However, \modelname{}-M5 encounters an performance saturation at 44.6 AP$^{box}$ and 40.6 AP$^{mask}$, albeit with a higher latency of 12.4ms compared to 9.3ms of RepNext-M5.

\begin{table*}[h]
\caption{\textbf{Ablation study over 120 epochs on ImageNet-1K}. 
Metrics reported include top-1 accuracy on the validation set, NPU latency on an iPhone 13 running iOS 18 (\textit{mlmodel}), CPU latency on an Intel(R) Xeon(R) Gold 6330 (\textit{onnx}), and throughput on a RTX3090 GPU.
Experiments were conducted at image resolutions of both 224 and 384 to validate the effectiveness of large receptive fields.
``\chunkmark'' represents the simultaneous recursive decomposition in both spatial and channel dimensions.
``$\circledast$'' indicates downsampling and upsampling operations are replaced by group convolutions.
``$\circleddash$'' denotes the element-wise addition is substituted by channel-wise concatenation.
``$\circledR$'' represents processing downsampled feature maps as sequential inputs through a recurrent architecture like RNN~\cite{650093} and SSM~\cite{gu2022efficiently,mamba}.
Nearest interpolation and max unpooling for upsampling can boost GPU throughput, but are not well supported by CoreML or ONNX.
}
\vspace{-4pt}
\centering
\small
\scalebox{0.95}{\input{tables/ablation}}
\label{tab:ablation}
\vspace{-10pt}
\end{table*}

\paragraph{Semantic Segmentation.}
We perform semantic segmentation experiments on the ADE20K dataset~\cite{zhou2017scene}, which consists of approximately 20K training images and 2K validation images across 150 categories.
Intersection-over-Union (mIoU) is used as the evaluation metric.
We followed the training protocol from previous works~\cite{li2022efficientformer,li2022rethinking} using the Semantic FPN framework~\cite{kirillov2019panoptic}. 
As illustrated in~\Cref{tab:coco}, \modelname{} demonstrates favorable mIoU-latency trade-offs across various model sizes, with \modelname{}-M3 achieving the highest mIoU of 41.0\% among other models of similar scale.
\modelname{}-M4 and \modelname{}-M5 also achieve competitive mIoU and latency compared to the state-of-the-art models.
These findings emphasize the critical role of efficient backbones with large receptive fields in advancing semantic segmentation. 

\subsection{Ablation Studies}
\vspace{-2pt}
We systematically investigate the impact of convolutional kernel sizes, recursive decomposition, upsampling operations, and \methodname{} variants on model performance across different image resolutions.
Experiments were conducted over 120 epochs on ImageNet-1K~\cite{deng2009imagenet} using \modelname{}-M1 as the baseline, at resolutions of $224 \times 224$ and $384 \times 384$.

\textbf{Kernel size.} Increasing convolution kernel sizes significantly enhance model performance, especially for higher-resolution images.
As shown in Table~\ref{tab:ablation}, expanding the kernel from $3 \times 3$ to $7 \times 7$ for $224 \times 224$ images boosts top-1 accuracy from 74.64\% to 75.52\% (+0.88\%). 
However, this increase comes with a rise in parameter count from 4.80M to 4.96M and a slight increase in latency from 1.24ms to 1.27ms.
Further enlarging the kernel to $11 \times 11$ slightly reduces accuracy to 75.41\%, while increasing parameters to 5.26M and rising latency to 1.31ms.
For $384 \times 384$ images, kernel enlargement delivers consistent accuracy improvements, achieving a +1.65\% enhancement (76.35\% → 78.00\%) as the kernel grows from $3 \times 3$ to $11 \times 11$, with a modest latency increase (+0.20ms).
These findings reveal a trade-off between the computational efficiency of smaller convolution kernels and the enhanced spatial feature capture of larger convolution kernels. 

\textbf{Recursive decomposition.} The integration of recursive decomposition expands the model's receptive field (7\arrowmark 80) and improves multi-scale feature aggregation, leading to accuracy gains across resolutions.
For $224\times 224$ images, decomposition with $5\times5$ kernels achieves a top-1 accuracy of 76.03\%, outperforming single-level configurations with larger kernel sizes of $7\times7$ (75.52\%) and $11\times11$ (75.41\%). 
The benefits are even more pronounced for larger image resolutions ($384\times384$), where the peak top-1 accuracy of 78.11\% is achieved compared to 77.42\%, without substantial changes in parameter count or FLOPs.
However, the method increases latency (2.2ms $\rightarrow$ 2.8ms) and significantly reduces throughput (1200 img/s $\rightarrow$ 300 img/s).
This is partly due to the widespread employment of bilinear upsampling, which could be substituted with transposed depthwise convolution, nearest interpolation, or max unpooling, albeit at the expense of reduced accuracy.
In summary, the progressive decomposition allows the network to effectively aggregate multi-frequency representations while balancing computational complexity by distributing operations across different spatial scales.

\textbf{Upsampling operation.} Table~\ref{tab:ablation} demonstrates the effectiveness of substituting bilinear upsampling with transposed depthwise convolution for $384\times384$ images,
While this replacement slightly increases computational complexity and parameter count, it potentially enhances feature fidelity during upsampling, resulting in a modest increase in top-1 accuracy from 78.03\% to 78.24\% with $7\times7$ kernels.
Additionally, it reduces latency from 2.84ms to 2.63ms and boosts throughput from 300img/s to 500img/s.
However, smaller kernel sizes (e.g., $5\times5$ and $3\times3$) lead to performance degradation, with accuracy dropping from 78.11\% to 77.87\% and from 77.72\% to 77.26\%, respectively.

\textbf{RecConv Variants.} We present \methodname{} variants incorporating recursive decomposition along both spatial and channel dimensions.
As summarized in Table~\ref{tab:ablation}, the variant with group convolutions reduces parameter from 5.81M to 5.76M, FLOPs from 2.75G to 2.70G, and increases throughput from 502 to 563, but results in a 0.33\% drop in accuracy (from 78.24\% to 77.91\%).
Meanwhile, the latency increases significantly due to the lack of adaptation of the transposed group convolution on the NPU.
The variant using channel-wise concatenation further reduces parameters by 0.4M and FLOPs by 0.1G, lowers latency by 0.3ms, and increases throughput by 240, but causes a more significant accuracy drop of 1.4\%.

\textbf{RecConv Aggregation.} 
We reformulate \methodname{}'s up-sampling aggregation (Eq. \ref{eq:recconv_ori}) to establish its connection to recurrent architectures like RNNs~\cite{650093} and SSMs~\cite{gu2022efficiently,mamba}.
This perspective views the downsampled feature maps across decomposition levels as a sequential input to a recurrent model (Eq. \ref{eq:recconv_rnn}).
Where $W^{l}$ represents the weight for the convolution of level $l$, while $W_{h}$ and $W_{x}$ are the weights of the hidden state and input, respectively.
$\uparrow$ denotes the spatial upsampling operation.
This modification slightly boosts accuracy but increases latency and FLOPs (detailed in Algorithm~\ref{alg:recconv_rnn} and Table~\ref{tab:ablation}).

\begin{equation}
  \label{eq:recconv_ori}
  h_l = W^{l} \ast \uparrow(h_{l-1}) + W^{l} \ast x_l
\end{equation}

\begin{equation}
  \label{eq:recconv_rnn}
  h_l = \uparrow(W_{h} \ast h_{l-1} + W_{x} \ast x_l)
\end{equation}

\begin{table}[h]
    \caption{\textbf{Ablation study over 300 epochs on ImageNet-1K at 256 image resolution with a batch size of 1024}. 
    Using two MLLA~\cite{han2024demystify} variants as baselines (gray).
    Metrics include top-1 validation accuracy, NPU latency on an iPhone 13 running iOS 18 (\textit{mlmodel}), CPU latency on an Intel(R) Xeon(R) Gold 6330 (\textit{onnx}), throughput on a RTX4090 GPU, and peak mixed-precision training memory usage. 
    ``$\dagger$'' denotes the application of linear attention with RecConv.
    ``$\ast$'' indicates partial CPU execution.
    }
    \vspace{-4pt}
    \centering
    \small
    \resizebox{0.48\textwidth}{!}{\input{tables/classification_mlla}}
    \label{tab:classification_mlla}
    \vspace{-10pt}
\end{table}

\textbf{RecConv Beyond.} We apply RecConv to MLLA variants, replacing linear attention and downsampling layers. 
As shown in Table~\ref{tab:classification_mlla}, modified models achieve comparable performance to baselines at similar FLOPs, with higher GPU throughput and reduced memory consumption.
RecConv offers performance advantages in smaller models but underperformed the baseline as parameter count increases, while maintaining faster inference and lower memory usage.
Models using nearest interpolation (Interp) for downsampling exhibit higher throughput and lower memory consumption compared to transposed depthwise convolution (Conv).
Employing linear attention with RecConv also achieves comparable performance to the baseline model while reducing FLOPs, GPU latency, and training memory.

\subsection{RecConv with Linear Attention}
\label{sec:recconv_linear_attention}

\begin{table}[h]
    \caption{\textbf{ImageNet-1K classification results.}
    The ``A'' series uses linear attention and nearest interpolation with RecConv, while the ``M'' series employs convolution and bilinear interpolation for simplicity and broader hardware compatibility (e.g., to address suboptimal nearest interpolation support in some iOS versions). 
    ``$\ast$'' refers to hard knowledge distillation.
    }
    \centering
    \small
    \resizebox{0.47\textwidth}{!}{\input{tables/classification_recnext}}
    \label{tab:classification_recnext}
  \vspace{-6pt}
\end{table}

\begin{figure}[t]
  \centering
  \includegraphics[width=0.95\linewidth]{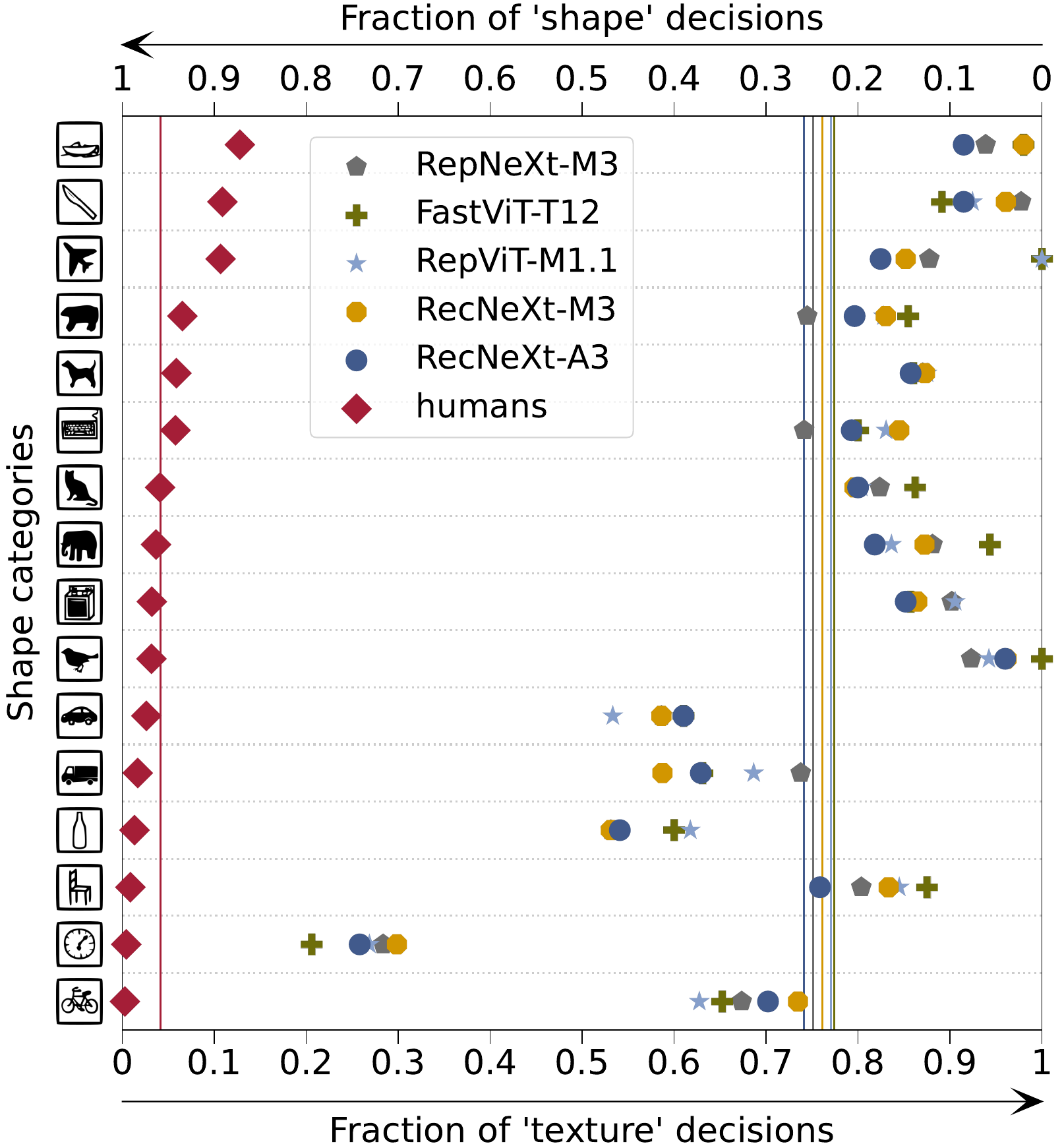}
  \caption{
    Shape bias comparison over 16 categories (\textit{modelvshuman}). 
    The vertical line is the average across categories.
  }
  \label{fig:shape-bias}
  \vspace{-6pt}
\end{figure}

\methodname{} is designed with great flexibility, enabling its convolutional modules to be easily replaced by alternative operations like linear attention~\cite{linear_attn,han2024demystify}.
As detailed in Table~\ref{tab:classification_recnext}, the RecNeXt models offer a compelling trade-off between accuracy, model size, and hardware-specific efficiency, achieving up to 83.1\% (83.5\% with knowledge distillation) Top-1 accuracy on ImageNet-1K with models from 2.5M to 25.7M parameters. 
The architecture provides two specialized paths: the M-series for accelerator-centric deployment with NPU latencies as low as 1.0ms, and the A-series for GPU-bound scenarios, where its linear attention and nearest interpolation design achieves higher throughput and better accuracy.

\begin{table}[h]
  \centering
  \small
  \caption{
  \textbf{Object detection and instance segmentation} were evaluated using Mask R-CNN on MS-COCO 2017. 
  \textbf{Semantic segmentation} results were obtained on ADE20K. Backbone FLOPs were measured with 512$\times$512 image resolution.
  }
  \resizebox{\linewidth}{!}{\input{tables/object_detection_recnext}}
  \label{tab:coco_recnext}
  \vspace{-6pt}
\end{table}

Table~\ref{tab:coco_recnext} presents a comparative analysis of our model families, evaluated across three distinct computational budgets on object detection, instance segmentation, and semantic segmentation tasks. 
At the most compact scale, the A3 model demonstrates superiority over M3 across all evaluated metrics while has fewer FLOPs.
As model complexity increases, a nuanced performance trade-off emerges. 
The M-series generally excels in object detection and instance segmentation. 
In contrast, the A-series consistently shows a competitive edge in semantic segmentation, with A3 achieving a notable +0.9 mIoU over M3 and A5 surpassing M5 by +0.5 mIoU. 

\subsection{Shape Bias Analysis}
\label{sec:shape_bias_analysis}

We evaluate shape bias—the fraction of predictions based on shape rather than texture—using the \textit{modelvshuman} benchmark~\cite{geirhos2021partial} in Figure~\ref{fig:shape-bias}. 
A higher shape bias aligns with human perception and is thus more desirable.
And we report quantitative results on the following datasets (Table~\ref{tab:model_performance}): Cue-Conflict (CC), Sketch (SK), High-Pass (HP), Low-Pass (LP), Silhouette (SI), Power-Equalisation (PE) and EidolonI (EI).
Our method outperforms other models across all 7 datasets while demonstrating superior shape bias.

\begin{table}[h]
  \centering
  \caption{Model Performance on 7 OOD Datasets}
  \label{tab:model_performance}
  \small
  \resizebox{0.48\textwidth}{!}{
  \begin{tabular}{lcccccccc}
  \toprule
  Model & FLOPs & CC ($\uparrow$) & SK ($\uparrow$) & HP ($\uparrow$) & LP ($\uparrow$) & SI ($\uparrow$) & PE ($\uparrow$) & EI ($\uparrow$) \\
  \midrule
  FastViT-T12 & 1.4 & 21.17 & 69.25 & 65.16 & 41.64 & 58.75 & 90.09 & 45.86 \\
  RepViT-M1.1 & 1.3 & 22.11 & 70.75 & 69.45 & 43.44 & 56.88 & 93.30 & 46.25 \\
  RepNeXt-M3  & 1.3 & 22.19 & 72.50 & 63.91 & 43.59 & 58.13 & 92.77 & 46.95 \\
  \rowcolor[HTML]{DAE8FC}
  RecNeXt-M3 & 1.4 & 22.11 & \textbf{73.50} & 66.02 & 42.73 & 58.75 & 93.30 & 46.95 \\
  \rowcolor[HTML]{DAE8FC}
  RecNeXt-A3 & 1.4 & \textbf{23.28} & 73.25 & \textbf{76.33} & \textbf{43.59} & \textbf{61.25} & \textbf{93.93} & \textbf{47.42} \\
  \bottomrule
  \end{tabular}
  }
\end{table}

\begin{figure*}[h]
  \centering
  \includegraphics[width=0.95\linewidth]{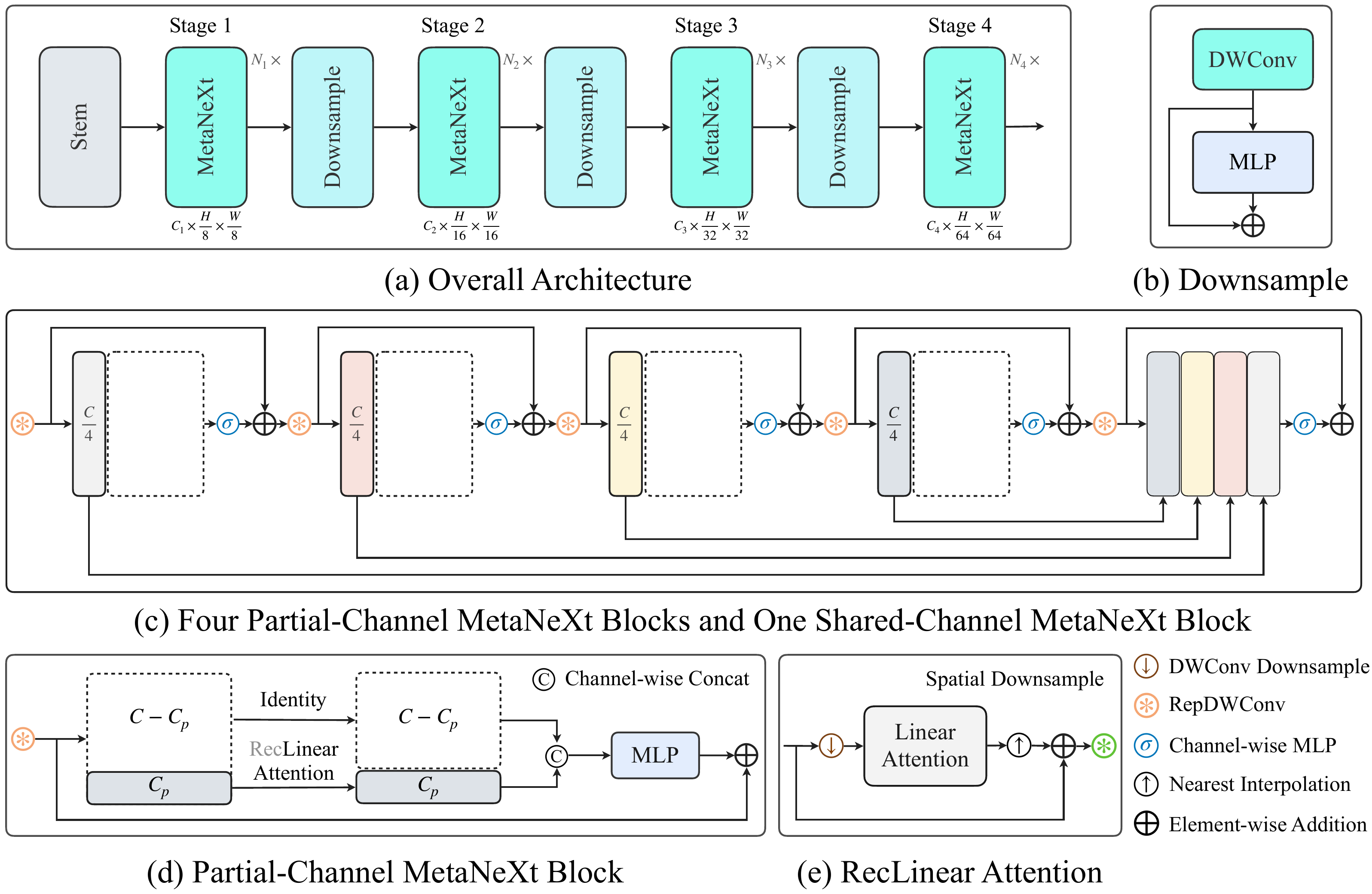}
  \caption{
    Architecture of the RecNeXt ``T'', ``S'', and ``B'' variants. 
    The overall architecture derives from LSNet~\cite{wang2025lsnetlargefocussmall}.
    The \textit{token mixer} of MetaNeXt~\cite{liu2022convnet,yu2023inceptionnext} implements partial channel operation~\cite{FasterNet, SHVIT} within the RecConv block using linear attention~\cite{linear_attn,han2024demystify}.
    To reduce computational cost and minimize feature redundancy, we employ shared channel operation in the final stage.
  }
  \label{fig:architecture}
\end{figure*}

\subsection{More Comparisons}
\label{sec:more_comparisons}

To further validate the effectiveness of \methodname{}, we compare it against the SOTA efficient architecture, LSNet~\cite{wang2025lsnetlargefocussmall} on ImageNet-1K.
We adopt the architecture design of LSNet, but replace the token mixer with partial channel~\cite{FasterNet, SHVIT} MetaNeXt~\cite{liu2022convnet,yu2023inceptionnext} style.
With shared channel operation in the final stage to further reduce computational cost and minimize feature redundancy, the detailed architecture is shown in Figure~\ref{fig:architecture}.
Where the ``RecLinear Attention'' denotes using linear attention in the \methodname{} block.

\begin{table}[h]
  \caption{
    \textbf{ImageNet-1K classification performance.}
    All models are trained for 300 epochs with a resolution of $224 \times 224$.
    Where ``$\dagger$'' denotes benchmark with Triton code.
    ``$\ast$'' refers to hard knowledge distillation.
    NPU latency is measured on an iPhone 13 with models compiled by Core ML Tools.
    CPU latency is accessed on a Quad-core ARM Cortex-A57 processor.
  }
  \centering
  \small
  \vspace{-7pt}
  \resizebox{0.47\textwidth}{!}{\input{tables/classification_lsnet}}
  \label{tab:classification_lsnet}
\end{table}

We evaluate ``T'', ``S'', ``B'' variants on ImageNet-1K as shown in Table~\ref{tab:classification_lsnet}.
The NPU latency is measured on an iPhone 13 with models compiled by Core ML Tools.
The CPU latency is accessed on a Quad-core ARM Cortex-A57 processor in ONNX format.
And the throughput is tested on a Nvidia RTX3090 with maximum power-of-two batch size that fits in memory.
``$\ast$'' denotes hard knowledge distillation using the RegNetY-16GF~\cite{radosavovic2020designing} as the teacher model.

The smallest model, RecNeXt-T, achieves a Top-1 accuracy of 75.2\%, which is a 0.3\% improvement over LSNet-T, with a negligible increase in parameters and identical FLOPs. 
With knowledge distillation, RecNeXt-T reaches 76.8\% accuracy, surpassing LSNet-T by 0.7\%. 
The model demonstrates impressive hardware efficiency, with an NPU latency of 1.8ms, an ARM CPU latency of 105ms, and a GPU throughput of 13,957 images per second.
RecNeXt-S achieves 78.3\% Top-1 accuracy, outperforming LSNet-S by 0.5\%, while having less parameters.
The distilled version of RecNeXt-S remains this gap, reaching 79.5\% accuracy.
RecNeXt-B matches the 80.3\% Top-1 accuracy of LSNet-B with fewer FLOPs and higher throughput. 
Across all model sizes, our architecture consistently delivers a competitive trade-off between accuracy, latency, and model size compared to LSNet. 
Table~\ref{tab:recnext_architecture} presents the detailed architecture, where the ``RecAttn Block'' denotes partial channel operation~\cite{FasterNet, SHVIT} of the RecConv block with linear attention.

%% file: tables/classification_without_distillation.tex
  \begin{tabular}{cccccccccc}
  \toprule
  \multirow{1}{*}{Model}          &  Latency & Params & Top-1 & \multicolumn{2}{c}{Auxiliary}
  \\
  & (ms) & (M) & (\%) & NAS & SRP
  \\
  \hline
  EfficientFormerV2-S0~\cite{li2022rethinking} & 1.4 & 3.5 & 73.7 & \cmark & \xmark \\
  FastViT-T8\textsuperscript{$\dagger$}~\cite{vasu2023fastvit} & 1.3 & 3.6 & 75.6 & \xmark & \cmark \\
  MobileOne-S1~\cite{vasu2023mobileone} & 1.2  & 4.8 & 75.9 & \xmark & \cmark \\
  RepViT-M0.9~\cite{wang2023repvit} & 1.3 & 5.1 & 77.4 & \xmark & \cmark \\
  RepNeXt-M1~\cite{zhao2024repnext} & 1.3 & 4.8 & 77.5 & \xmark & \cmark \\
  RepViT-M1.0~\cite{wang2023repvit} & 1.4 & 6.8 & 78.6 & \xmark & \cmark \\
  RepNeXt-M2~\cite{zhao2024repnext} & 1.4 & 6.5 & 78.9 & \xmark & \cmark \\
  \rowcolor[HTML]{ECF4FF}
  \textbf{\modelname{}-M1} & 1.4 & 5.2 & \textbf{78.0} & \xmark & \xmark \\
  \rowcolor[HTML]{DAE8FC}
  \textbf{\modelname{}-M2} & 1.5 & 6.8 & \textbf{79.2} & \xmark & \xmark \\
  \hline
  MobileOne-S2~\cite{vasu2023mobileone} & 1.5 & 7.8 & 77.4 & \xmark & \cmark \\
  EfficientFormerV2-S1~\cite{li2022rethinking} & 1.6 & 6.1 & 77.9 & \cmark & \xmark \\
  MobileOne-S3~\cite{vasu2023mobileone} & 1.7 & 10.1 & 78.1 & \xmark & \cmark \\
  FastViT-T12\textsuperscript{$\dagger$}~\cite{vasu2023fastvit} & 1.8 & 6.8 & 79.1 & \xmark & \cmark \\
  RepViT-M1.1~\cite{wang2023repvit} & 1.5 & 8.2 & 79.4 & \xmark & \cmark \\
  RepNeXt-M3~\cite{zhao2024repnext} & 1.5 & 7.8 & 79.4 & \xmark & \cmark \\
  \rowcolor[HTML]{DAE8FC}
  \textbf{\modelname{}-M3} & 1.6 & 8.2 & \textbf{79.6} & \xmark & \xmark \\
  \hline
  MobileOne-S4~\cite{vasu2023mobileone} & 2.2 & 14.8 & 79.4 & \xmark & \cmark \\
  FastViT-S12\textsuperscript{$\dagger$}~\cite{vasu2023fastvit} & 2.0 & 8.8 & 79.8 & \xmark & \cmark \\
  PoolFormer-S24~\cite{yu2022metaformer} & 3.2 & 21.0 & 80.3 & \xmark & \xmark \\
  EfficientFormerV2-S2~\cite{li2022rethinking} & 2.3 & 12.6 & 80.4 & \cmark & \xmark \\
  FastViT-SA12\textsuperscript{$\dagger$}~\cite{vasu2023fastvit} & 2.3 & 10.9 & 80.6 & \xmark & \cmark \\
  RepViT-M1.5~\cite{wang2023repvit} & 1.9 & 14.0 & 81.2 & \xmark & \cmark \\
  RepNeXt-M4~\cite{zhao2024repnext} & 1.9 & 13.3 & \text{81.2} & \xmark & \cmark \\
  \rowcolor[HTML]{DAE8FC}
  \textbf{\modelname{}-M4} & 2.4 & 14.1 & \textbf{81.4} & \xmark & \xmark \\
  \hline
  PoolFormer-S36~\cite{yu2022metaformer} & 4.3 & 31.0 & 81.4 & \xmark & \xmark \\
  RepViT-M2.3~\cite{wang2023repvit} & 2.8 & 22.9 & 82.5 & \xmark & \cmark \\
  FastViT-SA24\textsuperscript{$\dagger$}~\cite{vasu2023fastvit} & 3.4 & 20.6 & \text{82.6} & \xmark & \cmark \\
  RepNeXt-M5~\cite{zhao2024repnext} & 2.8 & 21.7 & 82.4 & \xmark & \cmark \\
  \rowcolor[HTML]{DAE8FC}
  \textbf{\modelname{}-M5} & 3.4 & 22.9 & \textbf{82.9} & \xmark & \xmark \\
  \bottomrule
  \end{tabular}

%% file: tables/object_detection.tex
\begin{tabular}{c|c|ccc|ccc|c}
\toprule
\multirow{2}{*}{Backbone} & \multirow{2}{*}{\makecell{Latency $\downarrow$ \\ (ms)}} & \multicolumn{3}{c|}{Object Detection} & \multicolumn{3}{c|}{Instance Segmentation} & \multicolumn{1}{c}{Semantic}  \\ \cline{3-9}
                          &                         & AP$^{box}$    & AP$^{box}_{50}$   & AP$^{box}_{75}$   & AP$^{mask}$    & AP$^{mask}_{50}$   & AP$^{mask}_{75}$   & mIoU   \\
                          \hline
                          \hline
ResNet18~\cite{he2016deep}                  &     3.8        & 34.0   & 54.0    & 36.7    & 31.2   & 51.0    & 32.7   & 32.9   \\
PoolFormer-S12~\cite{yu2022metaformer}            &     6.2          & 37.3   & 59.0    & 40.1    & 34.6   & 55.8    & 36.9     & 37.2  \\
FastViT-SA12~\cite{vasu2023fastvit}     &     7.6              &  \text{38.9}  &  \text{60.5}   & \text{42.2}  &    \text{35.9}    &   \text{57.6}      &  \text{38.1}   & \text{38.0}  \\
RepViT-M1.1~\cite{wang2023repvit}     &     4.1              &  \text{39.8}  &  \text{61.9}   & \text{43.5}  &    \text{37.2}    &   \text{58.8}      &  \text{40.1}   & \text{40.6}  \\
RepNeXt-M3~\cite{zhao2024repnext}     &     4.3              &  \text{40.8}  &  \text{62.4}   & \text{44.7}  &    \text{37.8}    &   \text{59.5}      &  \text{40.6}   & \text{40.6}  \\
\rowcolor[HTML]{DAE8FC}
\modelname{}-M3     &     \text{5.2}              &  \textbf{41.7}  &  \textbf{63.4}   & \textbf{45.4}  &    \textbf{38.6}    &   \textbf{60.5}      &  \textbf{41.4}   & \textbf{41.0}  \\
PoolFormer-S24~\cite{yu2022metaformer}            &     10.5            & 40.1   & 62.2    & 43.4    & 37.0   & 59.1    & 39.6  &  40.3 \\
PVT-Small~\cite{wang2021pyramid} & 26.2  & 40.4 & 62.9 & 43.8 & 37.8 & 60.1 & 40.3 & 39.8 \\
SwiftFormer-L1~\cite{Shaker_2023_ICCV}     &     6.2              &  \text{41.2}  &  \text{63.2}   & \text{44.8}  &    \text{38.1}    &   \text{60.2}      &  \text{40.7}   & \text{41.4}  \\
RepViT-M1.5~\cite{wang2023repvit}      &       5.7      & \text{41.6}   &  63.2  &  \text{45.3}   &  \text{38.6}   &  60.5   & \text{41.5}     &   \text{43.6}   \\
FastViT-SA24~\cite{vasu2023fastvit}    &       12.2              &  \text{42.0}  &  \text{63.5}   & \text{45.8}  &    \text{38.0}    &   \text{60.5}      &  \text{40.5}   & \text{41.0}  \\
RepNeXt-M4~\cite{zhao2024repnext}      &       6.1      & \text{42.9}   &  \text{64.4}  &  \text{47.2}   &  \text{39.1}   &  \text{61.7}   & \text{41.7}     &   \text{43.3}   \\
\rowcolor[HTML]{DAE8FC}
\modelname{}-M4      &       \text{7.6}      & \textbf{43.5}   &  \textbf{64.9}  &  \textbf{47.7}   &  \textbf{39.7}   &  \textbf{62.1}   & \textbf{42.4}     &   \textbf{43.6}   \\
SwiftFormer-L3~\cite{Shaker_2023_ICCV}     &     10.4              &  \text{42.7}  &  \text{64.4}   & \text{46.7}  &    \text{39.1}    &   \text{61.7}      &  \text{41.8}   & \text{43.9}  \\
FastViT-SA36~\cite{vasu2023fastvit}     &     16.2              &  \text{43.8}  &  \text{65.1}   & \text{47.9}  &    \text{39.4}    &   \text{62.0}      &  \text{42.3}   & \text{42.9}  \\
RepViT-M2.3\textsuperscript{$\dagger$}~\cite{wang2023repvit}      &       9.0      & 44.6   &  66.1  &  \text{48.8}   &  \textbf{40.8}   &  \textbf{63.6}  & \textbf{43.9}     &   \textbf{46.1}   \\
RepNeXt-M5~\cite{zhao2024repnext}      &       9.3      & \textbf{44.7}   &  66.0  &  \textbf{49.2}   &  \text{40.7}   &  \text{63.5}  & \text{43.6}     &   \text{45.0}   \\
\rowcolor[HTML]{DAE8FC}
\modelname{}-M5      &       \text{12.4}      & \text{44.6}   &  \textbf{66.3}  &  \text{49.0}   &  \text{40.6}   &  \text{63.5}  & \text{43.5}     &   \text{46.0}   \\
\bottomrule
\end{tabular}

%% file: tables/ablation.tex
\begin{tabular}{c | c | c c | c c c c c c | c } 
\toprule
  \multirow{2}{*}{\makecell[c]{Image Size}} & \multirow{2}{*}{\makecell[c]{Levels}}   & \multirow{2}{*}{\makecell[c]{Kernel Size}} &  \multirow{2}{*}{\makecell[c]{Receptive Field}}  & Params & MACs & \multicolumn{2}{c}{Latency (ms)} & Throughput & Top1 & Upsample \\
  ~ & ~ & ~ & ~ & (M) & (G) & NPU & CPU & (im/s) & (\%) & Operation \\
\hline
  \multirow{8}{*}{\makecell[c]{$224\times 224$}} & \multirow{3}{*}{\makecell[c]{-}} & $3\times3$ & [3, 3, 3, 3] & 4.80 & 0.82 & 1.24 & 7.4 & 4538 & 74.64 & \multirow{3}{*}{\makecell[c]{-}}  \\
  ~ & ~ & $7\times7$ & [7, 7, 7, 7] & 4.96 & 0.87 & 1.27 & 7.4 & 3583 & 75.52 & ~ \\
  ~ & ~ & $11\times11$ & [11, 11, 11, 11] & 5.26 & 0.96 & 1.31 & 7.5 & 2634 & 75.41 & ~ \\
\cdashline{2-11}
  ~ & \multirow{6}{*}{\makecell[c]{[4, 3, 2, 1]}} & $3\times3$ & [48, 24, 12, 6] & 4.90 & 0.83 & 1.33 & 12.9 & 389 & 75.48 & \multirow{3}{*}{\makecell[c]{Bilinear \\ Interpolate}} \\
  ~ & ~ & $5\times5$ & [80, 40, 20, 10] & 5.16 & 0.87 & 1.36 & 13.0 & 384 & \textbf{76.03} & ~ \\
  ~ & ~ & $7\times7$ & [112, 56, 28, 14] & 5.55 & 0.92 & 1.43 & 13.0 & 374 & 75.67 & ~ \\
\cdashline{3-11}
  ~ & ~ & ${\scriptsize \;}\ 7\times7^\circledR$ & [112, 56, 28, 14] & 5.72 & 0.99 & 1.46 & 17.2 & 2077 & 75.93 & \multirow{2}{*}{\makecell[c]{Nearest \\ Interpolate}} \\
  ~ & ~ & $7\times7$ & [112, 56, 28, 14] & 5.55 & 0.91 & 1.43 & 14.6 & 2677 & 75.71 & ~ \\
\cdashline{3-11}
  ~ & ~ & $7\times7$ & [112, 56, 28, 14] & 5.35 & 0.89 & - & - & 2770 & 75.48 & Max Unpool \\
\cline{0-10}
  \multirow{11}{*}{\makecell[c]{$384\times 384$}} & \multirow{3}{*}{\makecell[c]{-}} & $3\times3$ & [3, 3, 3, 3] & 4.80 & 2.40 & 2.07 & 13.2 & 1575 & 76.35  & \multirow{3}{*}{\makecell[c]{-}} \\
  ~ & ~ & $7\times7$ & [7, 7, 7, 7] & 4.96 & 2.55 & 2.20 & 13.2 & 1239 & 77.42 & ~ \\
  ~ & ~ & $11\times11$ & [11, 11, 11, 11] & 5.26 & 2.82 & 2.27 & 13.8 & 889 & 78.00 & ~ \\
\cdashline{2-11}
  ~ & \multirow{8}{*}{\makecell[c]{[4, 3, 2, 1]}} & $3\times3$ & [48, 24, 12, 6] & 4.90 & 2.44 & 2.71 & 20.6 & 320 & 77.72  & \multirow{3}{*}{\makecell[c]{Bilinear \\ Interpolate}} \\
  ~ & ~ & $5\times5$ & [80, 40, 20, 10] & 5.16 & 2.54 & 2.79 & 20.6 & 311 & \textbf{78.11} & ~ \\
  ~ & ~ & $7\times7$ & [112, 56, 28, 14] & 5.55 & 2.69 & 2.84 & 20.6 & 295 & 78.03 & ~ \\
\cline{3-11}
  ~ & ~ & $3\times3$ & [48, 24, 12, 6] & 4.97 & 2.44 & 2.40 & 41.5 & 1017 & \text{77.26} & \multirow{5}{*}{\makecell[c]{DWConv \\ Transpose}} \\
  ~ & ~ & $5\times5$ & [80, 40, 20, 10] & 5.31 & 2.57 & 2.48 & 63.5 & 616 & \text{77.87} & ~ \\
  ~ & ~ & $7\times7$ & [112, 56, 28, 14] & 5.81 & 2.75 & 2.63 & 94.1 & 502 & \textbf{78.24} & ~ \\
\cdashline{3-10}
  ~ & ~ & ${\scriptsize \text{\chunkmark}}\ 7\times7^\circledast$ & [112, 56, 28, 14] & 5.76 & 2.70 & - & 120.3 & 563 & 77.91 & ~ \\
  ~ & ~ & ${\scriptsize \text{\chunkmark}}\ 7\times7^\circleddash$ & [112, 56, 28, 14] & 5.43 & 2.64 & 2.33 & 54.3 & 740 & 76.83 & ~ \\
\bottomrule
\end{tabular}

%% file: tables/classification_mlla.tex
\begin{tabular}{cccccccccc} \toprule
Upsample & Params & FLOPs &  \multicolumn{2}{c}{Latency (ms)}  & Throughput & Memory & Top1
\\
Type & (M) & (G) & NPU\textsuperscript{$\ast$} & CPU & (im/s) & (G) & (\%) 
\\
\hline
\textcolor{gray}{-} & \textcolor{gray}{3.6} & \textcolor{gray}{0.9} & \textcolor{gray}{-} & \textcolor{gray}{-} &\textcolor{gray}{2601} & \textcolor{gray}{48.2} & \textcolor{gray}{76.3} \\
Conv & 4.1  & 0.9 & 34.7 & 26.9 & 2951 & 39.2 & 77.1 \\
Interp & 4.0  & 0.9 & 34.0 & 16.4 & 3196 & 38.7 & 77.1 \\
\;\;Interp\textsuperscript{$\dagger$} & 4.0  & 0.9 & 33.5 & 15.9 & 3282 & 39.1 & 77.3 \\
\hline
\textcolor{gray}{-} & \textcolor{gray}{13.9} & \textcolor{gray}{3.1} & \textcolor{gray}{-} & \textcolor{gray}{-} &\textcolor{gray}{1126} & \textcolor{gray}{97.3} & \textcolor{gray}{82.3} \\
Conv & 14.5 & 2.9 & 72.6 & 53.8 & 1408 & 79.8 & 82.1 \\
Interp & 14.3 & 2.9 & 70.8 & 31.9 & 1545 & 78.6 & 81.9 \\
\;\;Interp\textsuperscript{$\dagger$} & 14.4 & 2.8 & 71.6 & 31.9 & 1468 & 79.7 & 82.2 \\
\bottomrule
\end{tabular}

%% file: tables/classification_recnext.tex
\begin{tabular}{cccccccccc} \toprule
 & Params & FLOPs &  \multicolumn{2}{c}{Latency (ms)}  & Throughput & Top1
\\
 & (M) & (G) & NPU & CPU & (im/s) & (\%) 
\\
\hline
M0 & 2.5 & 0.4 & 1.0 & 189 & 750 & 73.2 / 74.7\textsuperscript{$\ast$} \\
A0 & 2.8 & 0.4 & 1.4 & 177 & 4891 & \textbf{73.6 / 75.0\textsuperscript{$\ast$}} \\
\hline
M1 & 5.2 & 0.9 & 1.4 & 361 & 384 & 78.0 / 79.2\textsuperscript{$\ast$} \\
A1 & 5.9 & 0.9 & 1.9 & 334 & 2730 & \textbf{78.3 / 79.6\textsuperscript{$\ast$}} \\
\hline
M2 & 6.8 & 1.2 & 1.5 & 431 & 325 & 79.2 / 80.3\textsuperscript{$\ast$} \\
A2 & 7.9 & 1.2 & 2.2 & 413 & 2331 & \textbf{79.6 / 80.8\textsuperscript{$\ast$}} \\
\hline
M3 & 8.2 & 1.4 & 1.6 & 482 & 314 & 79.6 / 80.9\textsuperscript{$\ast$} \\
A3 & 9.0 & 1.4 & 2.4 & 447 & 2151 & \textbf{80.1 / 81.1\textsuperscript{$\ast$}} \\
\hline
M4 & 14.1 & 2.4 & 2.4 & 843 & 169 & 81.4 / 82.5\textsuperscript{$\ast$} \\
A4 & 15.8 & 2.4 & 3.6 & 764 & 1265 & \textbf{81.6 / 82.5\textsuperscript{$\ast$}} \\
\hline
M5 & 22.9 & 4.7 & 3.4 & 1487 & 104 & 82.9 / 83.3\textsuperscript{$\ast$} \\
A5 & 25.7 & 4.7 & 5.6 & 1376 & 733 & \textbf{83.1 / 83.5\textsuperscript{$\ast$}} \\
\bottomrule
\end{tabular}

%% file: tables/object_detection_recnext.tex
\begin{tabular}{c|c|ccc|ccc|c}
\toprule
\multirow{2}{*}{} & \multirow{2}{*}{\makecell{FLOPs \\ (G)}} & \multicolumn{3}{c|}{Object Detection} & \multicolumn{3}{c|}{Instance Segmentation} & \multicolumn{1}{c}{Semantic}  \\ \cline{3-9}
                          &                         & AP$^{b}$    & AP$^{b}_{50}$   & AP$^{b}_{75}$   & AP$^{m}$    & AP$^{m}_{50}$   & AP$^{m}_{75}$   & mIoU   \\
                          \hline
                          \hline
M3 & \text{7.3}  & \text{41.7}   & \text{63.4}   & \text{45.4}   & \text{38.6}   & \text{60.5}   & \text{41.4}   & \text{41.0}   \\
A3 & \text{7.2}  & \textbf{42.1} & \textbf{64.1} & \textbf{46.2} & \textbf{38.8} & \textbf{61.2} & \textbf{41.6} & \textbf{41.9} \\
\hline
M4 & \text{12.4} & \text{43.5}   & \text{64.9}   & \textbf{47.7} & \text{39.7}   & \text{62.1}   & \text{42.4}   & \textbf{43.6} \\
A4 & \text{12.4} & \textbf{43.5} & \textbf{65.4} & \text{47.6}   & \textbf{39.8} & \textbf{62.4} & \textbf{42.9} & \text{43.0}   \\
\hline
M5 & \text{24.6} & \textbf{44.6} & \text{66.3}   & \textbf{49.0} & \textbf{40.6} & \textbf{63.5} & \textbf{43.5} & \text{46.0}   \\
A5 & \text{24.6} & \text{44.4}   & \textbf{66.3} & \text{48.9}   & \text{40.3}   & \text{63.3}   & \text{43.4}   & \textbf{46.5} \\
\bottomrule
\end{tabular}

%% file: tables/classification_lsnet.tex
\begin{tabular}{cccccccccc} \toprule
Model & Params & FLOPs &  \multicolumn{2}{c}{Latency (ms)}  & Throughput & Top1
\\
 & (M) & (G) & NPU & CPU & (im/s) & (\%) 
\\
\hline
LSNet-T~\cite{wang2025lsnetlargefocussmall} & 11.4 & 0.3 & - & 91 & 14708\textsuperscript{$\dagger$} & 74.9 / 76.1\textsuperscript{$\ast$} \\
RecNeXt-T & 12.1 & 0.3 & 1.8 & 105 & 13957\textsuperscript{$\dagger$} / 13301 & \textbf{75.2 / 76.8\textsuperscript{$\ast$}} \\
\hline
LSNet-S~\cite{wang2025lsnetlargefocussmall} & 16.1 & 0.5 & - & 139 & 9023\textsuperscript{$\dagger$} & 77.8 / 79.0\textsuperscript{$\ast$} \\
RecNeXt-S & 15.8 & 0.7 & 2.0 & 182 & 8034\textsuperscript{$\dagger$} / 7862 & \textbf{78.3 / 79.5\textsuperscript{$\ast$}} \\
\hline
LSNet-B~\cite{wang2025lsnetlargefocussmall} & 23.2 & 1.3 & - & 335 & 3996\textsuperscript{$\dagger$} & 80.3 / \textbf{81.6\textsuperscript{$\ast$}} \\
RecNeXt-B & 19.2 & 1.1 & 2.5 & 296 & 4472\textsuperscript{$\dagger$} / 4353 & \textbf{80.3} / 81.5\textsuperscript{$\ast$} \\
\bottomrule
\end{tabular}

%% file: sec/5_conclusion.tex
\section{Conclusions}

This paper introduces a simple yet flexible recursive decomposition strategy using small-kernel convolutions to construct multi-frequency representations, maintaining linear parameter growth with decomposition levels while ensuring computational complexity decreases exponentially (following geometric progression).
Leveraging this framework, we construct \methodname{} as a plug-and-play module seamlessly integrable into existing vision architectures.
To the best of our knowledge, this is the first attempt with effective receptive field up to $\mathbf{80 \times 80}$ for resource-constrained vision tasks.
Building on \methodname{}, we present \modelname{}, an efficient large-kernel vision backbone optimized towards resource-limited scenarios.
We introduce two series of models: the ``A'' series uses linear attention and nearest interpolation, while the ``M'' series employs convolution and bilinear interpolation for simplicity and broader hardware compatibility (e.g., to address sub-optimal nearest interpolation support in some iOS versions).
Extensive benchmarks demonstrate \modelname{}'s competitive performance over current leading approaches without requiring structural reparameterization or neural architecture search.

\textit{Limitations:}
The ``M'' series exhibits the lowest throughput among models of comparable parameter size due to extensive use of bilinear interpolation (mitigable via nearest interpolation, max unpooling or transposed convolution).
The recursive decomposition may introduce numerical instability during mixed precision training, which can be alleviated by using fixed-point or BFloat16 arithmetic.

\clearpage

%% file: sec/X_suppl.tex
\clearpage
\setcounter{page}{1}
\maketitlesupplementary

\section{RecConv Aggregation}
\label{sec:recconv_aggregation}

\begin{algorithm}[t]
\caption{Recurrent aggregation in a PyTorch-like style}
\label{alg:recconv_rnn}
\begin{lstlisting}[style=Python]
class RecConv(Module):
  def __init__(self, ic, ks=5, level=1):
      super().__init__()
      self.level = level
      kwargs = {
        'in_channels': ic,
        'out_channels': ic,
        'kernel_size': ks,
        'padding': ks // 2,
        'groups': ic
      }
      self.n = Conv(stride=2, **kwargs)
      self.a = Conv(**kwargs)
      self.b = Conv(**kwargs)
      self.c = Conv(**kwargs)
      self.d = Conv(**kwargs)

  def forward(self, x):
      fs = [x]
      for _ in range(self.level):
          fs.append(self.n(fs[-1]))
      # Multi-scale Recurrent Aggregation.
      h = None
      for i, o in reversed(zip(fs[1:], fs[:-1])):
          h = self.a(h) + self.b(i) if h else self.b(i)
          h = interpolate(h, size=o.shape[2:])
      return self.c(h) + self.d(x)
\end{lstlisting}
\end{algorithm}

Algorithm~\ref{alg:recconv_rnn} shows the implementation of the RecConv aggregation like SSM in a PyTorch-like style.
this approach treats the sequence of downsampled feature maps from each decomposition level as the input to a recurrent model.

\section{Grad-CAM Visualization}
\label{sec:gradcam_visualization}

\begin{figure}[h]
  \centering
  \includegraphics[width=0.95\linewidth]{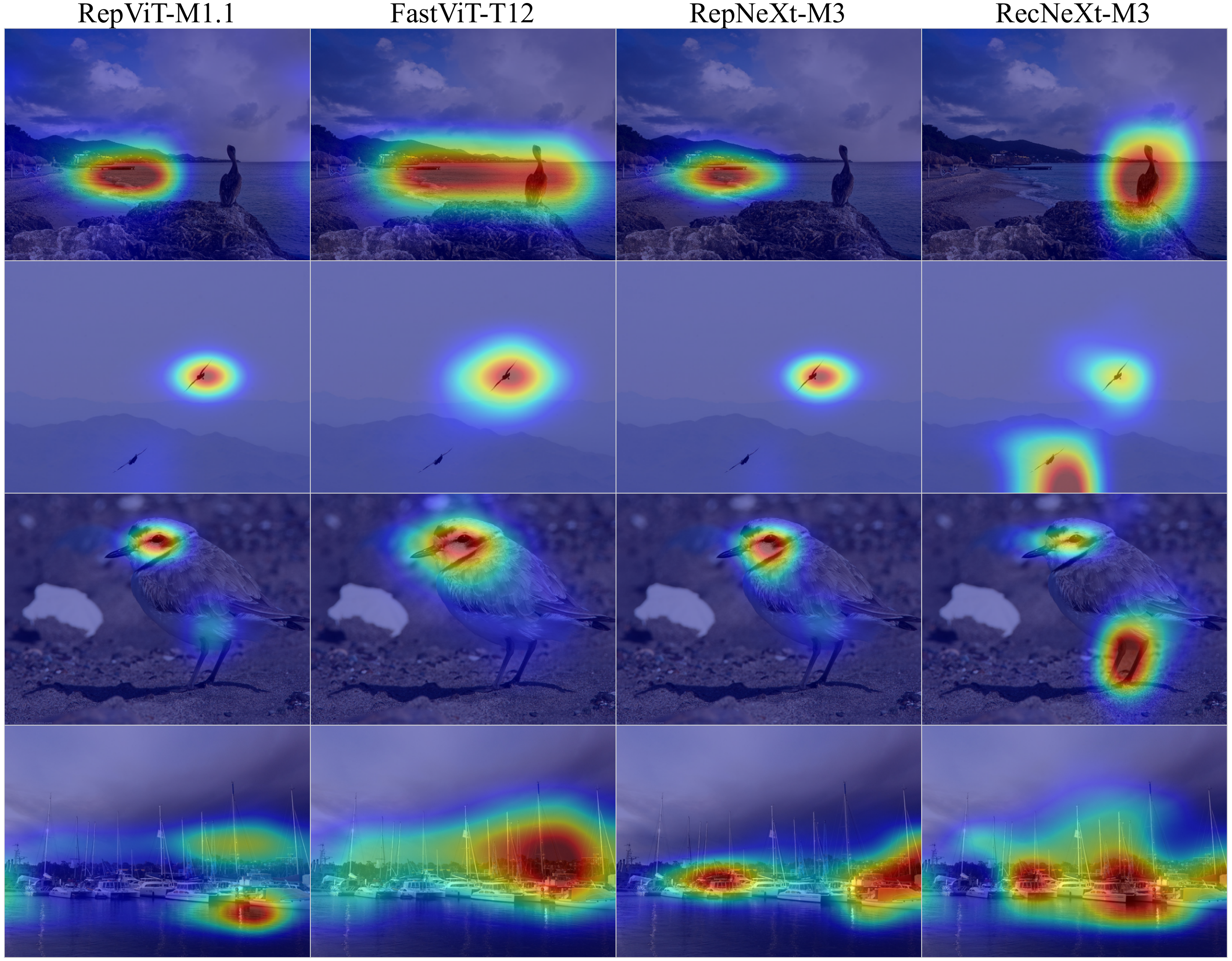}
  \caption{Grad-CAM~\cite{cvpr2017grad} on the MS-COCO validation dataset for RepViT-M1.1, FastViT-T12, RepNeXt-M3 and RecNeXt-M3.}
  \label{fig:gradcam_coco}
  \vspace{-10pt}
\end{figure}

We visualize class activation maps (CAM) using Grad-CAM~\cite{cvpr2017grad} with the TorchCAM Toolbox~\cite{torcham2020}.
As shown in Figure~\ref{fig:gradcam_coco}, \modelname{} achieves a better trade-off between local and global features compared to other efficient models.

\begin{table}[h]
  \caption{Architectural details of RecNeXt ``T'', ``S'', ``B'' variants.}
  \label{tab:recnext_architecture}
  \small
  \centering
  \resizebox{\columnwidth}{!}{%
  \begin{tabular}{c|c|c|c|c|c|c}
  \toprule
  \multirow{2}{*}{Stage} & \multirow{2}{*}{Resolution} &    \multirow{2}{*}{Type} & \multirow{2}{*}{Config} & \multicolumn{3}{c}{RecNeXt}  \\ \cline{5-7}
  & & & & T & S & B \\
  \hline
  \hline
  \multirow{6}{*}{stem} & \multirow{2}{*}{$\frac{H}{2} \times \frac{W}{2}$} & \multirow{2}{*}{Convolution} & \multirow{2}{*}{channels} & \multirow{2}{*}{16} & \multirow{2}{*}{32} & \multirow{2}{*}{32} \\
  & & & & & \\
  \cline{2-7}
  & \multirow{2}{*}{$\frac{H}{4} \times \frac{W}{4}$} & \multirow{2}{*}{Convolution} & \multirow{2}{*}{channels} & \multirow{2}{*}{32} & \multirow{2}{*}{64} & \multirow{2}{*}{64} \\
  & & & & & \\
  \cline{2-7}
  & \multirow{2}{*}{$\frac{H}{8} \times \frac{W}{8}$} & \multirow{2}{*}{Convolution} & \multirow{2}{*}{channels} & \multirow{2}{*}{64} & \multirow{2}{*}{128} & \multirow{2}{*}{128} \\
  & & & & & \\
  \hline
  \multirow{2}{*}{1} & \multirow{2}{*}{$\frac{H}{8} \times \frac{W}{8}$} & \multirow{2}{*}{RecAttn Block} & channels & 64 & 128 & 128 \\
  \cline{4-7}
  & & & blocks & 0 & 0 & 2 \\
  \hline
  \multirow{2}{*}{2} & \multirow{2}{*}{$\frac{H}{16} \times \frac{W}{16}$} & \multirow{2}{*}{RecAttn Block} & channels & 128 & 256 & 256 \\
  \cline{4-7}
  & & & blocks & 2 & 2 & 8 \\
  \hline
  \multirow{2}{*}{3} & \multirow{2}{*}{$\frac{H}{32} \times \frac{W}{32}$} & \multirow{2}{*}{RecAttn Block} & channels & 256 & 384 & 384 \\
  \cline{4-7}
  & &  & blocks & 8 & 8 & 8 \\
  \hline
  \multirow{2}{*}{4} & \multirow{2}{*}{$\frac{H}{64} \times \frac{W}{64}$} & \multirow{2}{*}{RecAttn Block} & channels & 512 & 512 & 512 \\
  \cline{4-7}
  & & & blocks & 10 & 10 & 12 \\
  \bottomrule
  \end{tabular}
  }
  \vspace{-10pt}
\end{table}

\section{Training Details}
\label{sec:training_details}

\begin{table}[h]
   \caption{
    Training recipe of ``M'' and ``A'' series and ``T'', ``S'', ``B'' variants of RecNeXt on ImageNet-1K.
    DropPath~\cite{larsson2016fractalnet} is set to $0.0$ for hard knowledge distillation.
  }
   \label{tab:recipe}
   \centering
   \small
   \begin{tabular}{cc}
   \toprule
   Hyperparameters & Config                        \\
   \hline
   optimizer       & AdamW                         \\
   batch size      & $1024$                        \\
   learning rate   & $2e-3$                        \\
   LR schedule     & cosine                        \\
   training epochs & $300$                         \\
   warmup epochs   & $5$                           \\
   weight decay    & $0.025$                       \\
   augmentation    & RandAug($9$, $0.5$)           \\
   color jitter    & $0.4$                         \\
   gradient clip   & $0.02$                        \\
   random erase    & $0.25$                        \\
   label smooth    & $0.1$                         \\
   mixup           & $0.8$                         \\
   cutmix          & $1.0$                         \\
  \multirow{2}{*}{drop path} & M4/A4: $0.2$ M5/A5: $0.3$ \\
  & T: $0.0$ S: $0.1$ B: $0.2$ \\
   \bottomrule
   \end{tabular}
\end{table}

We adopt the training recipe form RepViT~\cite{wang2023repvit} and LSNet~\cite{wang2025lsnetlargefocussmall} for ImageNet-1K classification.
All models are trained from scratch for 300 epochs using 224$\times$224 resolution input.
The initial learning rate is set to $2e-3$, and the total batch size is set to 1024. 
DropPath~\cite{larsson2016fractalnet} is applied to the larger variants for regularization without knowledge distillation.
Table~\ref{tab:recipe} provides the detailed training hyper-parameters.